\documentclass{nle}

\usepackage{amsmath}
\usepackage{graphicx}
\usepackage{caption}
\usepackage{subcaption}
\usepackage{multicol}
\usepackage{multirow}
\usepackage[T1]{fontenc}
\usepackage{natbib}
\usepackage{url}
\usepackage{csquotes}
\usepackage{xcolor}
\usepackage[ruled,vlined]{algorithm2e}
\ifpdf%
\usepackage{epstopdf}%
\else%
\fi

\begin{document}
\label{firstpage}

\lefttitle{Natural Language Engineering}
\righttitle{Yiping Jin et al.}

\papertitle{Article}

\jnlPage{1}{32}
\jnlDoiYr{2021}
\doival{10.1017/xxxxx}

\title{Toward Improving Coherence and Diversity of Slogan Generation}

\begin{authgrp}
\author{Yiping Jin$^1$}
\author{Akshay Bhatia$^2$}
\author{Dittaya Wanvarie$^{1,*}$}
\affiliation{$^1$Department of Mathematics and Computer Science, Faculty of Science,\\
         Chulalongkorn University, Bangkok, Thailand 10300\\
        \email{$^*$Corresponding author. Dittaya.W@chula.ac.th}}
\end{authgrp}

\begin{authgrp}
\author{ Phu T. V. Le$^2$}
\affiliation{$^2$Knorex, 140 Robinson Road, \#14-16 Crown @ Robinson, Singapore 068907\\
        \email{\{jinyiping, akshay.bhatia, le.phu\}@knorex.com}}
\end{authgrp}

\history{(Received 11 February 2021; revised 7 September 2021)}
%\received{20 March 1995; revised 30 September 1998}

\begin{abstract}
Previous work in slogan generation focused on utilising slogan skeletons mined from existing slogans. While some generated slogans can be catchy, they are often not coherent with the company's focus or style across their marketing communications because the skeletons are mined from other companies' slogans. We propose a sequence-to-sequence (seq2seq) transformer model to generate slogans from a brief company description. A na\"ive seq2seq model fine-tuned for slogan generation is prone to introducing false information. We use company name delexicalisation and entity masking to alleviate this problem and improve the generated slogans' quality and truthfulness. Furthermore, we apply conditional training based on the first words' POS tag to generate syntactically diverse slogans. Our best model achieved a ROUGE-1/-2/-L F$_1$ score of 35.58/18.47/33.32. Besides, automatic and human evaluations indicate that our method generates significantly more factual, diverse and catchy slogans than strong LSTM and transformer seq2seq baselines.
\end{abstract}

\maketitle

\section{Introduction}

Advertisements are created based on the market opportunities and product functions~\citep{white1972creativity}. Their purpose is to attract viewers' attention and encourage them to perform the desired action, such as going to the store or clicking the online ad. Slogans~\footnote{We use ``slogan'' and ``ad headline'' interchangeably. A \textit{slogan} is defined by its property as ``a short and memorable phrase used in advertising''. An \textit{ad headline} is defined literarily by its function.} are a key component in advertisements. Early studies in the fields of psychology and marketing revealed that successful slogans are \textbf{concise}~\citep{lucas1934optimum} and \textbf{creative}~\citep{white1972creativity}. Puns, metaphors, rhymes and proverbs are among the popular rhetorical devices employed in advertising headlines~\citep{phillips2009impact,mieder1977tradition}. However, as \citet{white1972creativity} noted, the creative process in advertising is ``within strict parameters''. I.e., the slogan must not diverge too much from the product/service it is advertising in its pursuit of creativity.

\begin{figure}
  \centering
  \includegraphics[width=.9\textwidth]{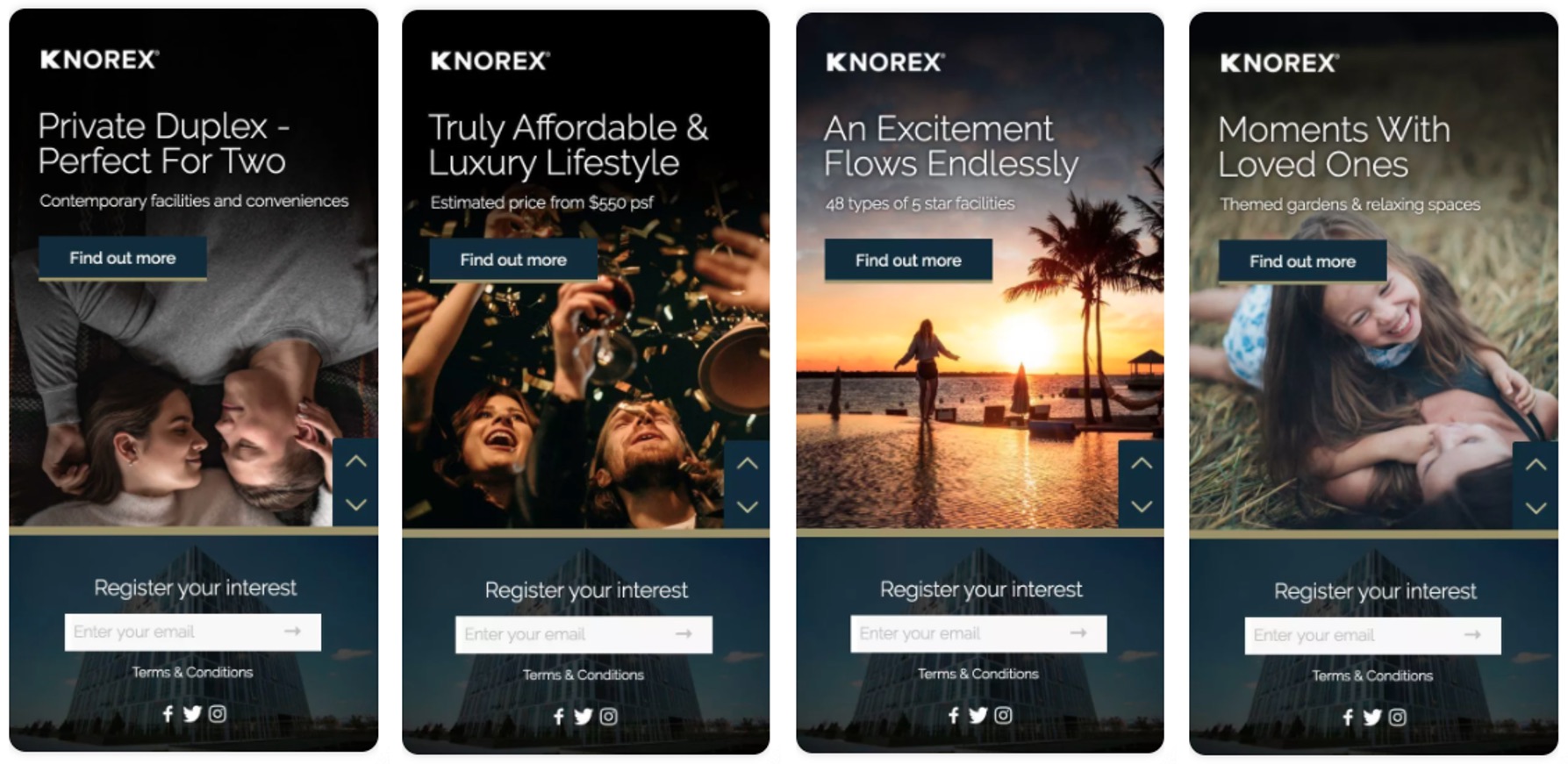}
\caption{Sample ads for the same advertiser in the hospitality industry. The centering text with the largest font corresponds to the ad headline (slogan).}
\label{fig:ad-example}
\end{figure}

Another essential factor to consider is ads fatigue~\citep{abrams2007personalized}. An ad's effectiveness decreases over time after users see it repeatedly. It motivates advertisers to deliver highly personalised and contextualised ads~\citep{vempati2020enabling}. While advertisers can easily provide a dozen alternative images and use different ad layouts to create new ads dynamically~\citep{bruce2017dynamic}, the ad headlines usually need to be manually composed. Figure~\ref{fig:ad-example} shows sample ads composed by professional ad creative designers, each having a different image and ad headline.

Previous work in automatic slogan generation focused almost exclusively on modifying existing slogans by replacing part of the slogan with new keywords or phrases~\citep{ozbal2013brainsup,tomavsic2014implementation,gatti2015slogans,alnajjar2020computational}. This approach ensures that the generated slogans are well-formed and attractive by relying on skeletons extracted from existing slogans. E.g., the skeleton ``The \textbf{NN} of Vintage'' expresses that the advertised product is elegant. It can instantiate novel slogans like ``The Phone of Vintage'' or ``The Car of Vintage''. However, the skeleton is selected based on the number of available slots during inference time and does not guarantee that it is coherent with the company or product. In this particular example, while some people appreciate vintage cars, the phrase ``The Phone of Vintage'' might have a negative connotation because it suggests the phone is \textit{outdated}. Such subtlety cannot be captured in skeletons represented either as part-of-speech tag sequences or syntactic parses.

In this work, we focus on improving \textbf{coherence} and \textbf{diversity} of a slogan generation system. We define coherence in two dimensions. First, the generated slogans should be consistent with the advertisers' online communication style and content. E.g., albeit being catchy, a pun is likely not an appropriate slogan for a personal injury law firm. To this end, we propose a sequence-to-sequence (seq2seq) transformer model to generate slogans from a brief company description instead of relying on random slogan skeletons. 

The second aspect of coherence is that the generated slogans should not contain untruthful information, such as mistaking the company's name or location. Therefore, we delexicalise the company name and mask entities in the input sequence to prevent the model from introducing unsupported information. 

Generating diverse slogans is crucial to avoid ads fatigue and enable personalisation. We observe that the majority of the slogans in our dataset are plain noun phrases that are not very catchy. It motivates us to explicitly control the syntactic structure through conditional training, which improves both diversity and catchiness of the slogans. 

We validate the effectiveness of the proposed method with both quantitative and qualitative evaluation. Our best model achieved a ROUGE-1/-2/-L F$_1$ score of 35.58/18.47/33.32. Besides, comprehensive evaluations also revealed that our proposed method generates more truthful, diverse, and catchy slogans than various baselines. The main contributions of this work are as follows:

\begin{itemize}
\setlength\itemsep{1em}
  \item Applying a transformer-based encoder-decoder model to generate slogans from a short company description.
  \item Proposing simple and effective approaches to improve the slogan's \textit{truthfulness}, focusing on reducing entity mention hallucination. 
  \item Proposing a novel technique to improve the slogan's syntactic \textit{diversity} through conditional training.
  \item Providing a benchmark dataset and a competitive baseline for future work to compare with.
\end{itemize}

We structure this paper as follows. We review related work on slogan generation, seq2seq models and aspects in generation in Section~\ref{sec:related-work}. In Section~\ref{sec:dataset}, we present the slogan dataset we constructed and conduct an in-depth data analysis. We describe our baseline model in Section~\ref{sec:model}, followed by our proposed methods to improve truthfulness and diversity in Section~\ref{sec:improve-factual} and Section~\ref{sec:syntactic-diversity}. We report the empirical evaluations in Section~\ref{sec:experiments}. Section~\ref{sec:ethical} presents ethical considerations and Section~\ref{sec:conclusion} concludes the paper and points directions for future work.

\section{Related Work}
\label{sec:related-work}
We review the literature in four related fields: (1) slogan generation, (2) sequence-to-sequence models, (3) truthfulness, and (4) diversity in language generation.

\subsection{Slogan Generation}

A slogan is a catchy, memorable, and concise message used in advertising. Traditionally, slogans are composed by human copywriters, and it requires in-depth domain knowledge and creativity. Previous work in automatic slogan generation mostly focused on manipulating existing slogans by injecting novel keywords or concepts while maintaining certain linguistic qualities. 

\cite{ozbal2013brainsup} proposed \textsc{B{\small RAIN}S{\small UP}}, the first framework for creative sentence generation that allows users to force certain words to be present in the final sentence and to specify various emotion, domain or linguistic properties.  \textsc{B{\small RAIN}S{\small UP}} generates novel sentences based on morpho-syntactic patterns automatically mined from a corpus of dependency-parsed sentences. The patterns serve as general skeletons of well-formed sentences. Each pattern contains several empty slots to be filled in. During generation, the algorithm first searches for the most frequent syntactic patterns compatible with the user's specification. It then fills in the slots using beam search and a scoring function that evaluates how well the user's specification is satisfied in each candidate utterance. 

\citet{tomavsic2014implementation} utilised similar slogan skeletons as \textsc{B{\small RAIN}S{\small UP}} capturing the POS tag and dependency type of each slot. Instead of letting the user specify the final slogan's properties explicitly, their algorithm takes a textual description of a company or a product as the input and parses for keywords and main entities automatically. They also replaced beam search with genetic algorithm to ensure good coverage of the search space. The initial population is generated from random skeletons. Each generation is evaluated using a list of ten heuristic-based scoring functions before producing a new generation using crossovers and mutations. Specifically, the mutation is performed by replacing a random word with another random word having the same POS tag. Crossover chooses a random pair of words in two slogans and switches them. E.g. input: [``Just \textit{do} it'', ``\textit{Drink} more milk''] $\Rightarrow$ [``Just \textit{drink} it'', ``\textit{Do} more milk'']. 

\citet{gatti2015slogans} proposed an approach to modify well-known expressions by injecting a novel concept from evolving news. They first extract the most salient keywords from the news and expand the keywords using WordNet and Freebase. When blending a keyword into well-known expressions, they check the word2vec embedding~\citep{mikolov2013distributed} similarity between each keyword and the phrase it shall replace to avoid generating nonsense output. \citet{gatti2015slogans} also used dependency statistics similar to \textsc{B{\small RAIN}S{\small UP}} to impose lexical and syntactic constraints. The final output is ranked by the mean rank of semantic similarity and dependency scores, thus balancing the relatedness and grammaticality. In subsequent work, \cite{gatti2017sing} applied a similar approach to modify song lyrics with characterizing words taken from daily news.

\citet{iwama2018japanese} presented a Japanese slogan generator using a slogan database, case frames, and word vectors. The system achieved an impressive result in an ad slogan competition for human copywriters and was employed by one of the world's largest advertising agencies. Unfortunately, their approach involves manually selecting the best slogans from ten times larger samples and they did not provide any detailed description of their approach. 

Recently, \cite{alnajjar2020computational} proposed a slogan generation system based on generating nominal metaphors. The input to the system is a target concept (e.g., car), and an adjective describing the target concept (e.g., elegant). The system generates slogans involving a metaphor such as ``The Car Of Stage'', suggesting that the car is as elegant as a stage performance. Their system extracts slogan skeletons from existing slogans. Given a target concept $T$ and a property $P$, the system identifies candidate metaphorical vehicles~\footnote{A metaphor has two parts: the tenor (target concept) and the vehicle. The vehicle is the object whose attributes are borrowed.} $v$. For each skeleton $s$ and the $\langle T,v \rangle$ pair, the system searches for potential slots that can be filled. After identifying plausible slots, the system synthesises candidate slogans optimised using genetic algorithms similar to \cite{tomavsic2014implementation}. 

\citet{munigala2018persuaide} is one of the pioneer works to use a language model (LM) to generate slogans instead of relying on slogan skeletons. Their system first identifies fashion-related keywords from the product specifications and expands them to creative phrases. They then synthesise persuasive descriptions from the keywords and phrases using a large domain-specific neural LM. Instead of letting the LM generate free-form text, the candidates at each time step are limited to extracted keywords, expanded in-domain noun phrases and verb phrases as well as common functional words. The LM minimises the overall perplexity with beam search. The generated sentence always begins with a \textit{verb} to form an imperative and persuasive sentence. \cite{munigala2018persuaide} demonstrated that their system produced better output than an end-to-end LSTM encoder-decoder model. However, the encoder-decoder was trained on a much smaller parallel corpus of title text-style tip pairs compared to the corpus they used to train the language model.

\cite{misawa2020distinctive} applied a Gated Recurrent Unit (GRU)~\citep{cho2014learning} encoder-decoder model to generate slogans from a discription of a target item. They argued that good slogans should not be generic but distinctive towards the target item. To enhance distinctiveness, they used a reconstruction loss~\citep{niu2019bi} by reconstructing the corresponding description from a slogan. They also employed a copying mechanism~\citep{see2017get} to handle out-of-vocabulary words occurring in the input sequence. Their proposed model achieved the best ROUGE-L score of 19.38~\footnote{The result was reported on a Japanese corpus. So it is not directly comparable to our work.}, outperforming various neural encoder-decoder baselines. 

Similarly, \cite{hughes2019generating} applied encoder-decoder with copying mechanism~\citep{see2017get} to generate search ad text from the landing page title and body text. They applied reinforcement learning (RL) to directly optimise for the click-through rate. \cite{mishra2020learning} also employed the same encoder-decoder model of \cite{see2017get} to ad text generation. However, their task is to rewrite a text with a low click-through rate to a text with a higher click-through rate (e.g., adding phrases like ``limited time offer'' or ``brand new'').

Concurrent to our work, \citet{kanungo2021ad} applied RL to a Transformer~\citep{vaswani2017attention} using the ROUGE-L score as the reward. Their model generates ad headlines from \textit{multiple} product titles in the same ad campaign. The generated headlines also need to generalise to multiple products instead of being specific to a single product. Their proposed method outperformed various LSTM and Transformer baselines based on overlap metrics and quality audits. Unfortunately, we could not compare with \citet{kanungo2021ad} because they used a large private dataset consisting of 500,000 ad campaigns created on Amazon. Their model training is also time expensive (over 20 days on an Nvidia V100 GPU).

Our approach is most similar to \cite{misawa2020distinctive} in that we also employ an encoder-decoder framework with a description as the input. However, we differ from their work in two principled ways. Firstly, we use a more modern Transformer architecture~\citep{vaswani2017attention}, which enjoys the benefit of extensive pre-training and outperforms recurrent neural networks in most language generation benchmarks. We do not encounter the problem of generating generic slogans and out-of-vocabulary words (due to subword tokenisation). Therefore, the model is greatly simplified and can be trained using a standard cross-entropy loss. Secondly, we propose simple yet effective approaches to improve the truthfulness and diversity of generated slogans.

\subsection{Sequence-to-Sequence Models}

\cite{sutskever2014sequence} presented a seminal sequence learning framework using multi-layer Long Short-Term Memory (LSTM)~\citep{hochreiter1997long}. The framework encodes the input sequence to a vector of fixed dimensionality, then decodes the target sequence based on the vector. This framework enables learning sequence-to-sequence (seq2seq)~\footnote{We use sequence-to-sequence and encoder-decoder interchangeably in this paper.} tasks where the input and target sequence are of a different length. \cite{sutskever2014sequence} demonstrated that their simple framework achieved close to state-of-the-art performance in an English to French translation task.

The main limitation of \cite{sutskever2014sequence} is that the performance degrades drastically when the input sequence becomes longer. It is because of unavoidable information loss when compressing the whole input sequence to a fixed-dimension vector. \cite{bahdanau2015neural} and \cite{luong2015effective} overcame this limitation by introducing attention mechanism to LSTM encoder-decoder. The model stores a contextualised vector for each time step in the input sequence. During decoding, the decoder computes the attention weights dynamically to focus on different contextualised vectors. Attention mechanism overtook the previous state-of-the-art in English-French and English-German translation and yields much more robust performance for longer input sequences.

LSTM, or more generally recurrent neural networks, cannot be fully parallelised on modern GPU hardware because of an inherent temporal dependency. The hidden states need to be computed one step at a time. \cite{vaswani2017attention} proposed a new architecture, the Transformer, which is based solely on multi-head self-attention and feed-forward layers. They also add positional encodings to the input embeddings to allow the model to use the sequence's order. The model achieved a new state-of-the-art performance, albeit taking a much shorter time to train than LSTM with attention mechanism.

\cite{devlin2019bert} argued that the standard Transformer~\citep{vaswani2017attention} suffers from the limitation that they are unidirectional and every token can only attend to previous tokens in the self-attention layers. To this end, they introduced BERT, a pre-trained bidirectional transformer by using a masked language model (MLM) pre-training objective. MLM masks some random tokens with a {\fontfamily{qcr}\selectfont
[MASK]} token and provides a bidirectional context for predicting the masked tokens. Besides, \cite{devlin2019bert} used the next sentence prediction task as an additional pre-training objective.

Despite achieving state-of-the-art results on multiple language understanding tasks, BERT does not make predictions auto-regressively, reducing its effectiveness for generation tasks. \cite{lewis-etal-2020-bart} presented BART, a model combining a bidirectional encoder (similar to BERT) and an auto-regressive decoder. This combination allows BART to capture rich bidirectional contextual representation, and yield strong performance in language generation tasks. Besides MLM, \cite{lewis-etal-2020-bart} introduced new pre-training objectives, including masking text spans, token deletion, sentence permutation, and document rotation. These tasks are particularly suitable for a seq2seq model like BART because there is no one-to-one correspondence between the input and target tokens.

\cite{zhang2020pegasus} employed an encoder-decoder transformer architecture similar to BART. They introduced a novel pre-training objective specifically designed for abstractive summarisation. Instead of masking single tokens (like BERT) or text spans (like BART), they mask whole sentences (referred to as ``gap sentences'') and try to reconstruct these sentences from their context. \cite{zhang2020pegasus} demonstrated that the model performs best when using important sentences selected greedily based on the ROUGE-F$_1$ score between the selected sentences and the remaining sentences. Their proposed model PEGASUS achieved state-of-the-art performance on all 12 summarisation tasks they evaluated. It also performed surprisingly well on a low-resource setting due to the relatedness of the pre-training task and abstractive summarisation.

While large-scale transformer-based language models demonstrate impressive text generation capabilities, users cannot easily control particular aspects of the generated text. \cite{keskarCTRL2019} proposed CTRL, a conditional transformer language model conditioned on control codes that influence the style and content. Control codes indicate the domain of the data, such as Wikipedia, Amazon reviews and subreddits focusing on different topics. \cite{keskarCTRL2019} use naturally occurring words as control codes and prepend them to the raw text prompt. Formally, given a sequence of the form $x = (x_1, . . . , x_n)$ and a control code $c$, CTRL learns the conditional probability $p_{\theta}(x_i \vert x_{<i}, c)$. By changing or mixing control codes, CTRL can generate novel text with very different style and content.

In this work, we use BART model architecture due to its flexibility as a seq2seq model and competitive performance on language generation tasks. We were also inspired by CTRL and applied a similar idea to generate slogans conditioned on additional attributes.

\subsection{Truthfulness in Language Generation}
\label{subsec:literature-factual}

While advanced seq2seq models can generate realistic text resembling human-written ones, they are usually optimised using a token-level cross-entropy loss. Researchers observed that a low training loss or a high ROUGE score do not guarantee the generated text is truthful with the source text~\citep{cao2018faithful,scialom2019answers}. Following previous work, we define \textit{truthfulness} as the generated text can be verified through the source text without any external knowledge.

We did not find any literature specifically addressing truthfulness in slogan generation. Most prior work investigated abstractive summarisation because 1) truthfulness is critical in the summarisation task, and 2) the abstractive nature encourages the model to pull information from different parts of the source document and fuse them. Therefore, abstractive models are more prone to hallucination compared to extractive models~\citep{durmus2020feqa}. Like summarisation, slogan generation also requires the generated content to be truthful. A prospect may feel annoyed or even be harmed by false information in advertising messages. 

Prior work focused mostly on devising new metrics to measure the truthfulness between the source and generated text. Textual entailment (aka. natural language inference) is closely related to truthfulness. If a generated sequence can be inferred from the source text, it is likely to be truthful. Researchers have used textual entailment to rerank the generated sequences~\citep{falke2019ranking,maynez2020faithfulness} or remove hallucination from the training dataset~\citep{matsumaru2020improving}. \citet{pagnoni2021understanding} recently conducted a comprehensive benchmark on a large number of truthfulness evaluation metrics and concluded that entailment-based approaches yield the highest correlation with human judgement. 

Another direction is to extract and represent the fact explicitly using information extraction techniques. \citet{goodrich2019assessing} and \citet{zhu2021enhancing} extracted relation tuples using OpenIE~\citep{angeli2015leveraging} while \citet{zhang2020optimizing} used a domain-specific information extraction system for radiology reports. The truthfulness is then measured by calculating the overlap between the information extracted from the source and generated text.

In addition, researchers also employed QA-based approaches to measure the truthfulness of summaries. \citet{eyal2019question} generated slot-filling questions from the \textit{source document} and measured how many of these questions can be answered from the generated summary. \citet{scialom2019answers} optimised towards a similar QA-based metric directly using reinforcement learning and demonstrated that it generated summaries with better relevance. Conversely, \citet{durmus2020feqa} and \citet{wang2020asking} generated natural language questions from the \textit{system-output summary} using a seq2seq question generation model and verified if the answers obtained from the source document agree with the answers from the summary.

Most recently, some work explored automatically correcting factual inconsistencies from the generated text. For example, \citet{dong2020multi} proposed a model-agnostic post-processing model that either iteratively or auto-regressively replaces entities to ensure semantic consistency. Their approach predicts a text span in the source text to replace an inconsistent entity in the generated summary. Similarly, \citet{chenimproving} modelled factual correction as a classification task. Namely, they predict the most plausible entity in the source text to replace each entity in the generated text that does not occur in the source text. 

Our work is most similar to \citet{dong2020multi} and \citet{chenimproving}. However, their methods require performing additional predictions on each entity in the generated text using BERT. It drastically increases the latency. We decide on a much simpler approach of replacing each entity in both the source and target text with a unique mask token before training the model, preventing it from generating hallucinated entities in the first place. We can then perform a trivial dictionary lookup to replace the mask tokens with their original surface form.

\subsection{Diversity in Language Generation}
\label{subsec:literature-diversity}

Neural language models often surprisingly generate bland and repetitive output despite their impressive capability, a phenomenon referred to as neural text degeneration~\citep{holtzman2019curious}. \citet{holtzman2019curious} pointed out that maximising the output sequence's probability is ``unnatural''. Instead, humans regularly use vocabulary in the low probability region, making the sentences less dull. While beam search and its variations~\citep{reddy1977speech,li2016simple} improved over greedy encoding by considering multiple candidate sequences, they are still maximising the output probability by nature, and the candidates often differ very little from each other. 

A common approach to improve diversity and quality of generation is to introduce randomness by sampling~\citep{ackley1985learning}. Instead of always choosing the most likely token(s) at each time step, the decoding algorithm samples from the probability distribution over the whole vocabulary. The shape of the distribution can be controlled using the temperature parameter. Setting the temperature to (0,1) shifts the probability mass towards the more likely tokens. Lowering the temperature improves the generation quality at the cost of decreasing diversity~\citep{caccia2019language}.

More recently, top $k$-sampling~\citep{fan2018hierarchical} and nucleus sampling~\citep{holtzman2019curious} were introduced to truncate the candidates before performing the sampling. Top $k$-sampling samples from a fixed most probable $k$ candidate tokens while nucleus (or top $p$) sampling samples from the most probable tokens whose probability sum is at least $p$. Nucleus sampling can dynamically adjust the top-$p$ vocabulary size. When the probability distribution is flat, the top-$p$ vocabulary size is larger, and when the distribution is peaked, the top-$p$ vocabulary size is smaller. \citet{holtzman2019curious} demonstrated that nucleus sampling outperformed various decoding strategies, including top-$k$ sampling. Besides, the algorithm can generate text that matches the human perplexity by tuning the threshold $p$.

\citet{welleck2019neural} argued that degeneration is not only caused by the decoding algorithm but also due to the use of maximum likelihood training loss. Therefore, they introduced an additional unlikelihood training loss. Specifically, they penalise the model for generating words in previous context tokens and sequences containing repeating n-grams. The unlikelihood training enabled their model to achieve comparable performance as nucleus sampling using only greedy decoding.

Instead of relying on randomness, we generate syntactically diverse slogans with conditional training similar to CTRL~\citep{keskarCTRL2019}. Automatic and human evaluations confirmed that our method yields more diverse and interesting slogans than nucleus sampling.

\section{Datasets}
\label{sec:dataset}

While large advertising agencies might have conducted hundreds of thousands of ad campaigns and have access to the historical ads with slogans~\citep{kanungo2021ad}, such a dataset is not available to the research community. Neither is it likely to be released in the future due to data privacy concerns. 

On the other hand, online slogan databases such as Textart.ru~\footnote{\tt http://www.textart.ru/database/slogan/list-advertising-slogans.html} and Slogans Hub~\footnote{\tt https://sloganshub.org/} contain at most hundreds to thousands of slogans, which are too few to form a training dataset, especially for a general slogan generator not limited to a particular domain. Besides, these databases do not contain company descriptions. Some even provide a list of slogans without specifying their corresponding company or product. They might be used to train a language model producing slogan-like utterance~\citep{gpt2slogan}, but it will not be of much practical use because we do not have control over the generated slogan's content. 

We observe that many company websites use their company name plus their slogan as the HTML page title. Examples are ``Skype $\vert$ Communication tool for free calls and chat'' and ``Virgin Active Health Clubs - Live Happily Ever Active''. Besides, many companies also provide a brief description in the ``description'' field in the HTML $<$meta$>$ tag~\footnote{\tt https://www.w3schools.com/tags/tag\_meta.asp}. Therefore, our model's input and output sequence can potentially be crawled from company websites.

We crawl the title and description field in the HTML $<$meta$>$ tag using the Beautiful Soup library~\footnote{\tt https://www.crummy.com/software/BeautifulSoup/} from the company URLs in the Kaggle 7+ Million Company Dataset~\footnote{\tt  https://www.kaggle.com/peopledatalabssf/free-7-million-company-dataset/}. The dataset provides additional fields, but we utilise only the company name and URL in this work. The crawling took around 45 days to complete using a cloud instance with two vCPUs. Out of the 7M companies, we could crawl both the $<$meta$>$ tag description and the page title for 1.4M companies. This dataset contains much noise due to the apparent reason that not all companies include their slogan in their HTML page title. We perform various cleaning/filtering steps based on various keywords, lexicographical and semantic rules. The procedure is detailed in Appendix~\ref{apx:data-cleaning}.

After all the cleaning and filtering steps, the total number of (description, slogan) pairs is 340k, at least two orders of magnitude larger than any publicly available slogan database. We reserve roughly 2\% of the dataset for validation and test each. The remaining 96\% is used for training (328k pairs). The validation set contains 5,412 pairs. For the test set, the first author of this paper manually curated the first 1,467 company slogans in the test set, resulting in 1,000 plausible slogans (68.2\%). The most frequent cases he filtered out are unattractive ``slogans'' with a long list of products/services, such as ``Managed IT Services, Network Security, Disaster Recovery'', followed by the cases where HTML titles containing alternative company names that failed to be delexicalised and some other noisy content such as address. We publish our validation and manually-curated test dataset for future comparisons~\footnote{\tt https://github.com/YipingNUS/slogan-generation-dataset}.

\begin{figure}
\centering
\begin{subfigure}{.5\textwidth}
  \centering
  \includegraphics[width=.95\textwidth]{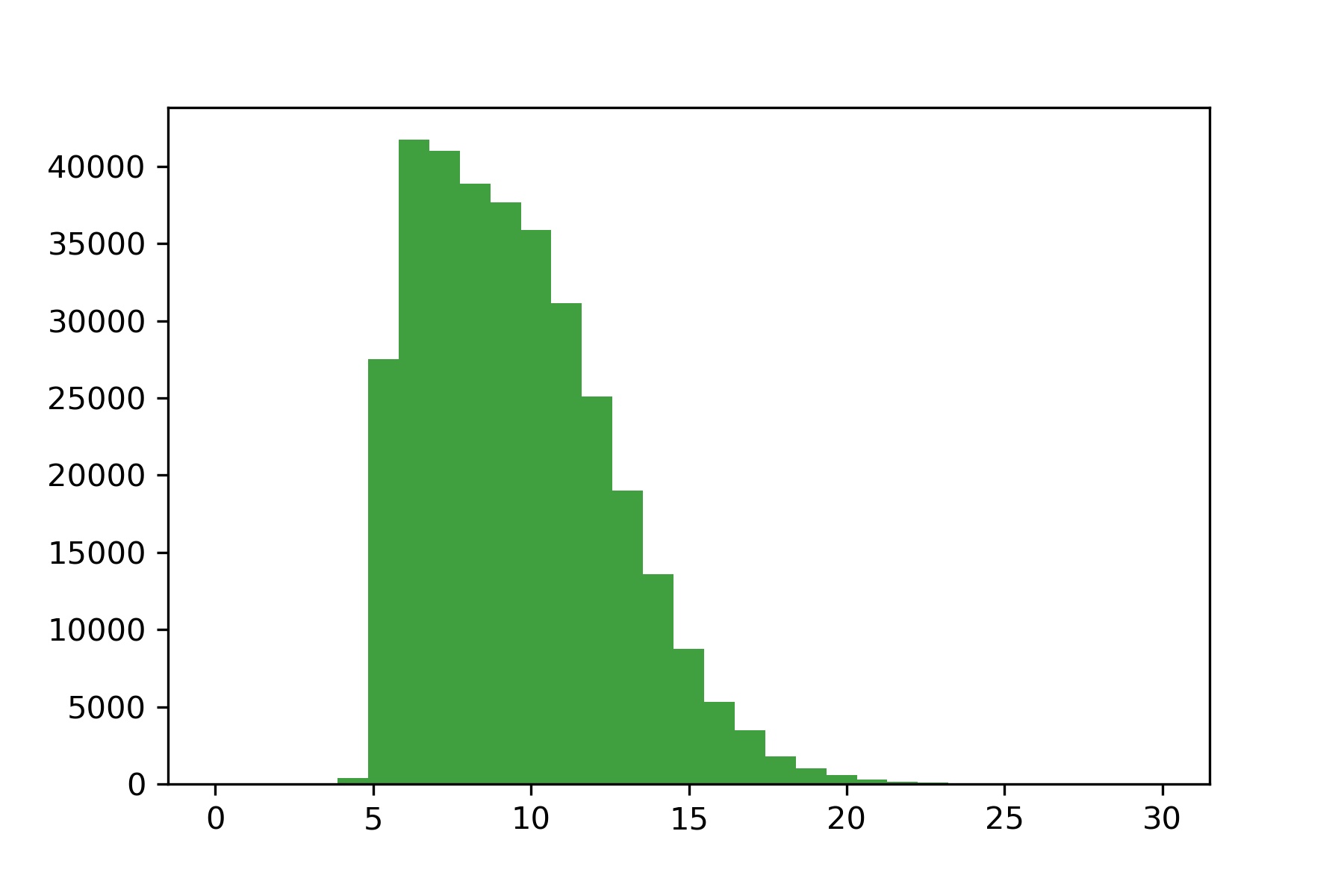}
  \caption{Slogans}
  \label{fig:slogan_len_hist}
\end{subfigure}%
\begin{subfigure}{.5\textwidth}
  \centering
  \includegraphics[width=.95\textwidth]{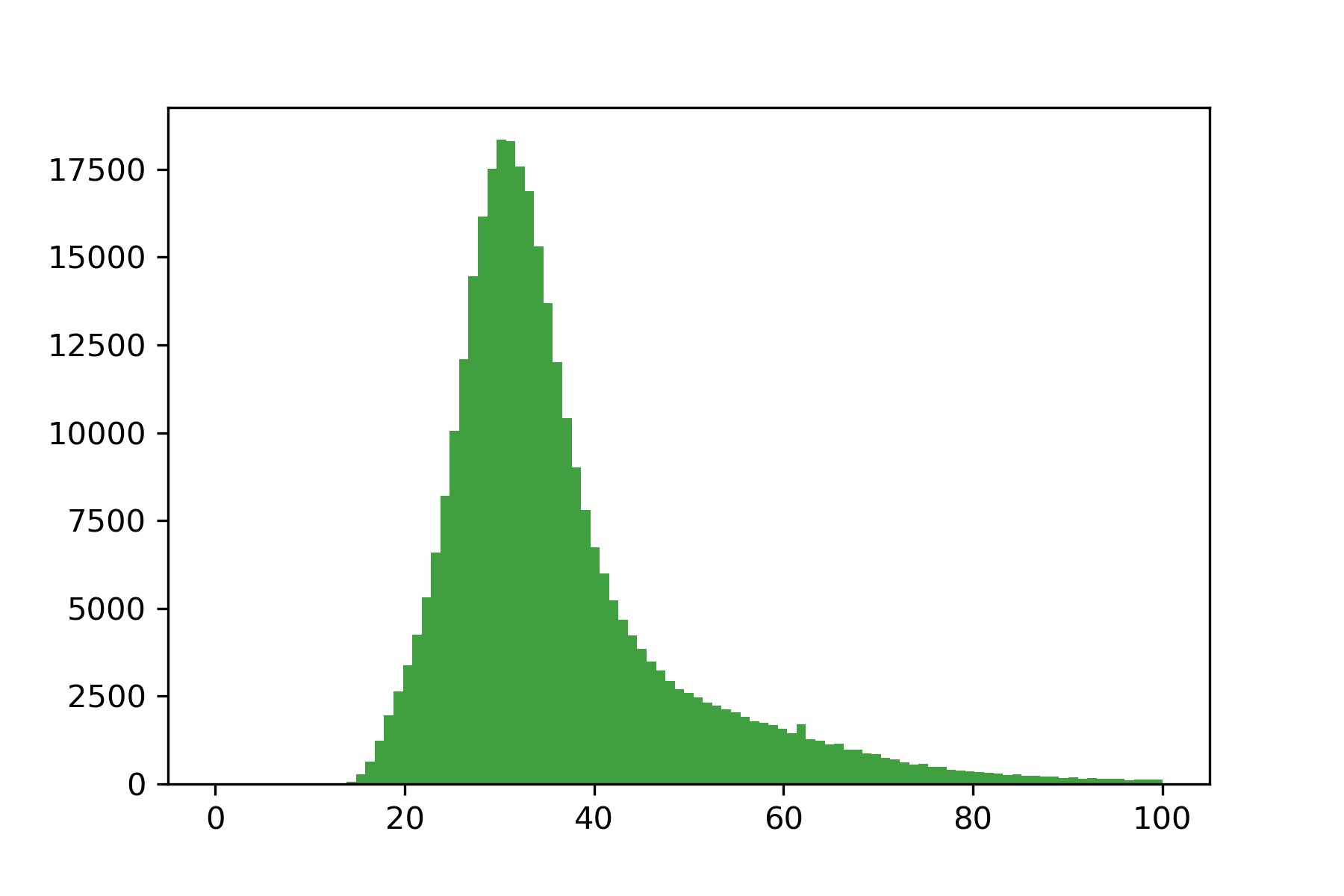}
  \caption{Descriptions}
  \label{fig:desc_len_hist}
\end{subfigure}
\caption{Distribution of the number of tokens in (a) slogans, and (b) descriptions.}
\label{fig:token_len_hist}
\end{figure}

We perform some data analysis on the training dataset to better understand the data. We first tokenise the dataset with BART's subword tokeniser. Figure~\ref{fig:token_len_hist} shows the distribution of the number of tokens in slogans and descriptions. While the sequence length of the description is approximately normally distributed, the length of slogans is right-skewed. It is expected because slogans are usually concise and contain few words. We choose a maximum sequence length of 80 for the description and 20 for the slogan based on the distribution. 

\begin{figure}
  \centering
  \includegraphics[width=.5\textwidth]{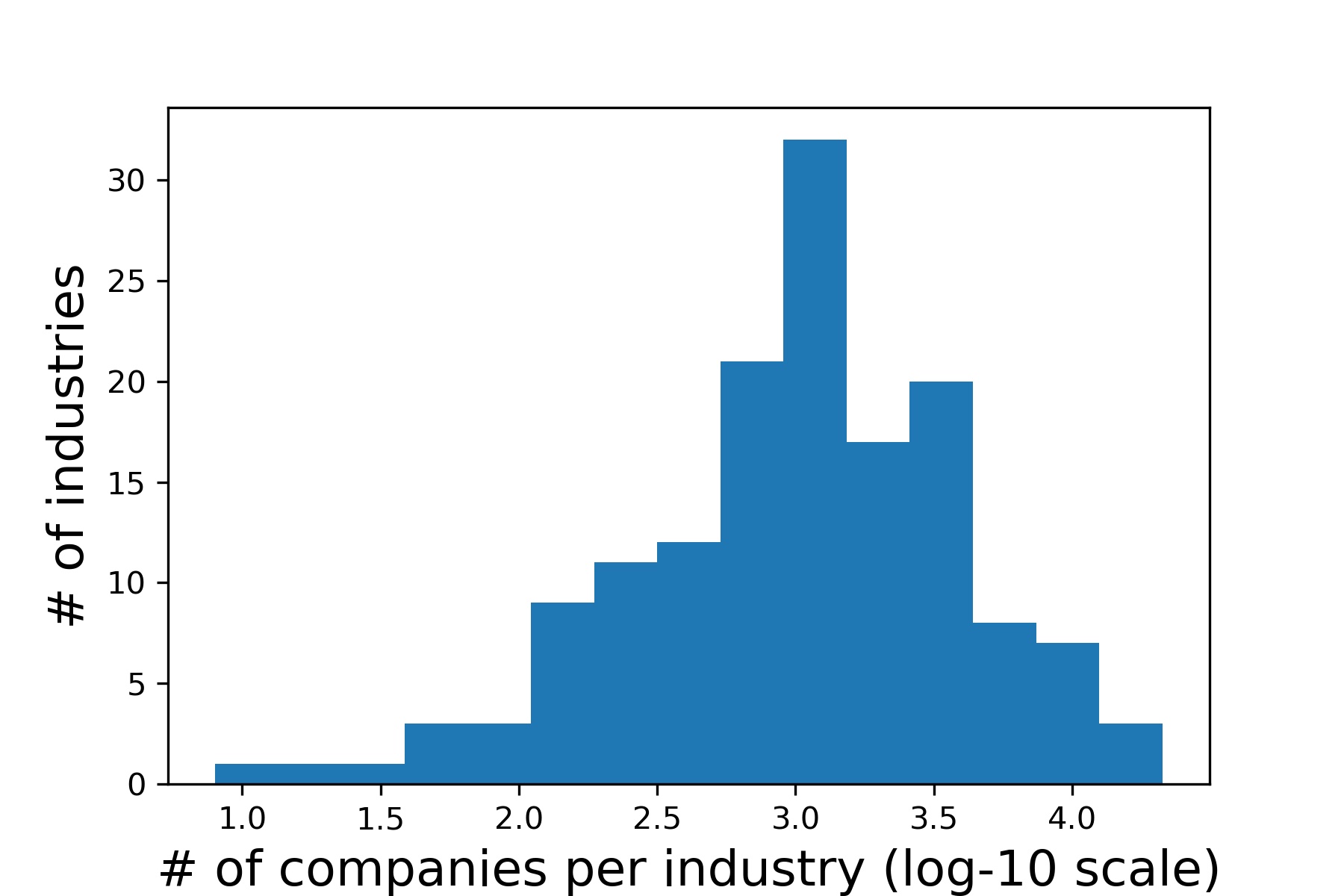}
\caption{Distribution of the number of companies belonging to each industry in log-10 scale.}
\label{fig:industry_hist}
\end{figure}

The training dataset covers companies from 149 unique industries (based on the ``industry'' field in the Kaggle dataset). Figure~\ref{fig:industry_hist} shows the distribution of the number of companies belonging to each industry on a log-10 scale. As we can see, most industries contain between $10^2$ (100) and $10^{3.5}$ (3,162) companies. Table~\ref{tab:top-industries} shows the most frequent ten industries with the number of companies and the percentage in the dataset. The large number of industries suggests that a model trained on the dataset will have observed diverse input and likely generalise to unseen companies.

\begin{table}[h!]
  \centering
  \caption{The most frequent ten industries in the training dataset.}
    \begin{tabular}{lrc}
    \hline
   	\textbf{Industry} & \textbf{\# of Companies} & \textbf{\%} \\\hline
	Information Technology and Services & 21,149 & 6.4\% \\\hdashline
	Marketing and Advertising & 15,691 & 4.8\% \\\hdashline
	Construction & 12,863 & 3.9\% \\\hdashline
	Computer Software & 11,367 & 3.5\% \\\hdashline
	Real Estate & 10,207 & 3.1\% \\\hdashline
	Internet & 10,025 & 3.1\% \\\hdashline
	Health, Wellness and Fitness & 9,513 & 2.9\% \\\hdashline
	Financial Services & 8,480 & 2.6\% \\\hdashline
	Automotive & 8,351 & 2.5\% \\\hdashline
	Retail & 8,217 & 2.5\% \\
    \hline
    \end{tabular}
  \label{tab:top-industries}
\end{table}

Furthermore, we investigate the following questions to understand the nature and the abstractness of the task:

\begin{enumerate}
\setlength\itemsep{0.5em}
  \item How many per cent of the slogans can be generated using a purely extractive approach, i.e., the slogan is contained in the description?
  \item How many per cent of the unigram words in the slogans occur in the description?
  \item How many per cent of the descriptions contain the company name? (We removed the company name from all the slogans).
  \item How many per cent of the slogans and descriptions contain entities? What are the entity types?
  \item How many per cent of the entities in the slogans do not appear in the description.
  \item Is there any quantitative difference between the validation and manually-curated test set that makes either of them more challenging?
\end{enumerate}

First, 11.2\% of the slogans in both the validation and the test set are contained in the descriptions (we ignore the case when performing substring matching). It indicates that approximately 90\% of the slogans require different degrees of abstraction. On average, 62.7\% of the word unigrams in the validation set slogans are contained in their corresponding descriptions, while the percentage for the test set is 59.0\%. 

63.1\% and 66.6\% of the descriptions in the validation and test set contain the company name. It shows that companies tend to include their name in the description, and there is an opportunity for us to tap on this regularity. 

We use Stanza~\citep{qi2020stanza} fine-grained named entity tagger with 18 entity types to tag all entities in the descriptions and slogans. Table~\ref{tab:entity-stats} presents the percentage of text containing each type of entity~\footnote{Details of the entity types can be found in the Ontonotes documentation: \url{https://catalog.ldc.upenn.edu/docs/LDC2013T19/OntoNotes-Release-5.0.pdf}.}. Besides {\fontfamily{qcr}\selectfont ORGANIZATION}, the most frequent entity types are {\fontfamily{qcr}\selectfont GPE}, {\fontfamily{qcr}\selectfont DATE}, {\fontfamily{qcr}\selectfont CARDINAL}, {\fontfamily{qcr}\selectfont LOCATION}, and {\fontfamily{qcr}\selectfont PERSON}. Many entities in the slogans do not appear in the corresponding description. It suggests that training a seq2seq model using the dataset will likely encourage entity hallucinations, which are commonly observed in abstractive summarisation. We show sample (description, slogan) pairs belonging to different cases in Table~\ref{tab:sample-slogans}. 

\begin{table}[h!]
  \centering
  \caption{The percentage of descriptions and slogans containing each type of entity. ``Slog\,-\,Desc'' indicates entities in the slogan that are not present in the corresponding description.}
    \begin{tabular}{lrrrrrr}
    \hline
   	&\multicolumn{3}{c}{Valid Dataset} & \multicolumn{3}{c}{Test Dataset}  \\
    Entity Type & Desc & Slogan & Slog\,-\,Desc & Desc & Slogan & Slog\,-\,Desc \\\hline
	ORGANIZATION & 65.6 & 31.3 & 27.8 & 63.9 & 30.2 & 29.7 \\\hdashline
	GPE & 36.7 & 19.4 & 7.5 & 33.5 & 20.2 & 7.5 \\\hdashline
	DATE & 16.4 & 1.3 & 1.0 & 18.2 & 1.9 & 1.4 \\\hdashline
	CARDINAL & 10.2 & 1.4 & 0.8 & 10.4 & 1.1 & 0.6 \\\hdashline
	LOCATION & 4.6 & 1.1 & 0.6 & 4.6 & 1.1 & 0.7 \\\hdashline
	PERSON & 4.2 & 2.5 & 1.6 & 3.3 & 1.3 & 0.9 \\\hdashline
	PRODUCT & 4.2 & 0.2 & 0.1 & 4.2 & 0.4 & 0.4 \\\hdashline
	NORP & 2.6 & 0.9 & 0.4 & 3.8 & 0.6 & 0.1 \\\hdashline
	FACILITY & 2.5 & 0.5 & 0.4 & 2.9 & 0.1 & 0.1 \\\hdashline
	TIME & 2.0 & 0.02 & - & 1.4 & - & - \\\hdashline
	WORK OF ART & 1.5 & 0.4 & 0.3 & 1.7 & 0.6 & 0.5 \\\hdashline
	PERCENT & 1.3 & 0.09 & 0.09 & 1.9 & - & - \\\hdashline
	ORDINAL & 1.3 & 0.2 & 0.1 & 1.4 & 0.3 & 0.2 \\\hdashline
	MONEY & 0.7 & 0.2 & 0.1 & 0.8 & - & - \\\hdashline
	QUANTITY & 0.5 & - & - & 0.4 & 0.1 & -\\\hdashline
	EVENT & 0.5 & 0.2 & 0.2 & 0.3 & 0.8 & 0.8 \\\hdashline
	LAW & 0.3 & - & - & 0.3 & - & -\\\hdashline
	LANGUAGE & 0.3 & 0.09 & 0.02 & 0.2 & 0.1 & - \\\hline
    \end{tabular}
  \label{tab:entity-stats}
\end{table}

\begin{table}[h!]
\centering
  \caption{Sample (description, slogan) pairs belonging to different cases from the validation set. We highlight the exact match words in bold.}
    \begin{tabular}{p{2cm}p{4cm}p{6cm}}
    \hline
   	\textbf{Remark} & \textbf{Slogan} & \textbf{Description} \\\hdashline
    Slogan in desc & Total Rewards Software & Market Total Rewards to Employees and Candidates with \textbf{Total Rewards Software}. We provide engaging Total Compensation Statements and Candidate Recruitment. \\\hdashline
    100\% unigrams in desc & Algebra Problem Solver & Free math \textbf{problem solver} answers your \textbf{algebra} homework questions with step-by-step explanations. \\\hdashline
    57\% unigrams in desc & Most Powerful Lead Generation Software for Marketers & \textbf{Powerful lead generation software} that converts abandoning visitors into subscribers with our dynamic marketing tools and Exit Intent technology. \\\hdashline
    33\% unigrams in desc & Business Process Automation & We help companies become more efficient by automating processes, impact \textbf{business} outcomes with actionable insights \& capitalize on growth with smart applications. \\\hdashline
    0\% unigrams in desc & Build World-Class Recreation Programs & Easily deliver personalized activities that enrich the lives of residents in older adult communities. Save time and increase satisfaction. \\\hdashline
    Entities in desc & Digital Agency in \textbf{Auckland$_{[GPE]}$} \& \textbf{Wellington$_{[GPE]}$}, New Zealand & Catch Design is an independent digital agency in \textbf{Auckland} and \textbf{Wellington}, NZ. We solve business problems through the fusion of design thinking, creativity, innovation and technology. \\\hdashline
    Entities not in desc & Leading Corporate Advisory Services Provider in \textbf{Singapore$_{[GPE]}$} \& \textbf{Hong Kong$_{[GPE]}$} & Offers Compliance Advisory services for Public listed companies, Private companies, NGOs, Offshore companies, and Limited Liability Partnerships (LLPs).
    \\\hline
    \end{tabular}
  \label{tab:sample-slogans}
\end{table}

The only notable difference that might make the test dataset more challenging is that it contains a slightly higher percentage of unigram words not occurring in the description than the validation dataset (41\% vs 37.3\%). However, this difference is relatively small, and we believe the performance measured on the validation dataset is a reliable reference when a hand-curated dataset is not available.

\section{Model}
\label{sec:model}

We apply a Transformer-based sequence-to-sequence (seq2seq) model to generate slogans. The model's input is a short company description. We choose BART encoder-decoder model~\citep{lewis-etal-2020-bart} with a bidirectional encoder and an autoregressive (left-to-right) decoder. BART enjoys the benefit of capturing bidirectional context representation like BERT and is particularly strong in language generation tasks.

We use DistilBART~\footnote{\tt https://huggingface.co/sshleifer/distilbart-cnn-6-6} with 6 layers of encoders and decoders each and 230M parameters. The model was a distilled version of BART$_{LARGE}$ trained by the HuggingFace team, and its architecture is equivalent to BART$_{BASE}$. We choose this relatively small model to balance generation quality and latency because our application requires generating multiple variations of slogans in real-time in a web-based user interface. 

The seq2seq slogan generation from the corresponding description is analogous to abstractive summarisation. Therefore, we initialise the model's weights from a \textit{fine-tuned} summarisation model on the CNN/DailyMail dataset~\citep{hermann2015teaching} instead of from a pre-trained model using unsupervised learning objectives. We freeze up to the second last encoder layer (including the embedding layer) and fine-tune the last encoder layer and the decoder. We do not fine-tune the whole model for two reasons: 1) we do not want the model to unlearn its ability to perform abstractive summarisation 2) by freezing a large portion of the parameters, we require much less RAM and can train using a much larger batch size.

\section{Generating Truthful Slogans}
\label{sec:improve-factual}

As we highlighted in Section~\ref{subsec:literature-factual}, generating slogans containing false or extraneous information is a severe problem for automatic slogan generation systems. In this section, we propose two approaches to improve the quality and truthfulness of generated slogans, namely delexicalising company names~(Section~\ref{subsec:method:delexicalisation}) and masking named entities~(Section~\ref{subsec:factual-entity}).

\subsection{Company Name Delexicalisation}
\label{subsec:method:delexicalisation}

Slogans should be concise and not contain extraneous information. Although we removed the company names from all slogans during preprocessing (described in Appendix~\ref{apx:data-cleaning}), we observe that a baseline seq2seq model often copies the company name from the description to the slogan. Table~\ref{tab:sample-extraneous-information} shows two such example generated by the seq2seq model. Both examples seem to be purely extractive except for changing the case to title case. The second example seems especially repetitive and is not a plausible slogan. As shown in Section~\ref{sec:dataset}, over 60\% of the descriptions contain the company name. Therefore, a method is necessary to tackle this problem. 

\begin{table}[h!]
\centering
  \caption{Examples of generated slogans containing the company name.}
    \begin{tabular}{lp{10cm}}
    \hline
   	\textbf{Company Name:} & Eftpos Warehouse \\
    \textbf{Description:} & 	
The latest \underline{Eftpos Warehouse} \& Point of Sale tech for the lowest prices. Flexible monthly rental options with the backing of our dedicated, on-call support team. \\
    \textbf{Generated Slogan:} & \underline{Eftpos Warehouse} \& Point of Sale Tech \\\hdashline
    \textbf{Company Name:} & MCB Financial Services \\
    \textbf{Description:} & 	
Financial Advisers Norwich, Norfolk - \underline{MCB Financial Services} Norwich are committed to helping you with your financial needs. \\
    \textbf{Generated Slogan:} & Financial Advisers Norwich, Norfolk - \underline{MCB Financial Services} Norwich \\
    \hline
    \end{tabular}
  \label{tab:sample-extraneous-information}
\end{table}

We apply a simple treatment to prevent the model from generating slogans containing the company name - delexicalising company name mentions in the description and replacing their surface text with a generic mask token {\fontfamily{qcr}\selectfont<company>}. After the model generates a slogan, any mask token is substituted with the original surface text~\footnote{Since the {\fontfamily{qcr}\selectfont<company>} token never occurs in slogans in our dataset, we have not observed a single case where the model generates a sequence containing the {\fontfamily{qcr}\selectfont<company>} token. We include the substitution for generality.}. 

We hypothesise that delexicalisation helps the model in two ways. Firstly, it helps the model avoid generating the company name by masking it in the input sequence. Secondly, the mask token makes it easier for the model to focus on the surrounding context and pick salient information to generate slogans.

The company name is readily available in our system because it is required when any new advertiser registers for an account. However, we notice that companies often use their shortened names instead of their official/legal name. Examples are ``Google LLC'' almost exclusively referred to as ``Google'' and ``Prudential Assurance Company Singapore (Pte) Limited'' often referred to as ``Prudential''. Therefore we replace the longest prefix word sequence of the company name occurring in the description with a {\fontfamily{qcr}\selectfont<company>} mask token. The process is illustrated in Algorithm~\ref{alg:delexilisation} (we omit the details handling the case and punctuations in the company name for simplicity). 

Besides the delexicalised text, the algorithm also returns the surface text of the delexicalised company name, which will replace the mask token during inference. It is also possible to use a more sophisticated delexicalisation approach, such as relying on a knowledge base or company directory such as Crunchbase to find alternative company names. However, the simple substitution algorithm suffices our use case. Table~\ref{tab:delexicalisation-example} shows an example description before and after delexicalisation.

\begin{algorithm}[H]
\SetAlgoLined
\KwIn{company\_name, text, MASK\_TOKEN}
\KwResult{delexicalised\_text, surface\_form}
 delexicalised\_text = text\;
 surface\_form = company\_name + `` ''\;
 \While{surface\_form.contains(`` '')}{
  surface\_form = surface\_form.substring(0, surface\_form.lastIndexOf(`` ''))\; 
  \If{text.contains(surface\_form)}{
   delexicalised\_text = text.replace(surface\_form, MASK\_TOKEN)\;
   break\;
   }
 }
 \caption{Prefix matching for delexicalising company names.}
 \label{alg:delexilisation}
\end{algorithm}

\begin{table}[h!]
\centering
  \caption{An example description before and after performing delexicalisation.}
    \begin{tabular}{p{2cm}p{10cm}}
    \hline
   	\textbf{Company:} & 
Atlassian Corporation Plc \\\hdashline
    \textbf{Description:} & 	
Millions of users globally rely on \textbf{Atlassian} products every day for improving software development, project management, collaboration, and code quality.\\
    \hdashline
     \textbf{Surface Form:} & Atlassian\\ 
    \textbf{Delexicalised Description:} & Millions of users globally rely on {\fontfamily{qcr}\selectfont<company>} products every day for improving software development, project management, collaboration, and code quality.\\
    \hline
    \end{tabular}
  \label{tab:delexicalisation-example}
\end{table}

\subsection{Entity Masking}
\label{subsec:factual-entity}

Introducing irrelevant entities is a more challenging problem compared to including company names in the slogan. It has been referred to as entity hallucination in the abstractive summarisation literature~\citep{nan-etal-2021-entity}. In a recent human study, \citet{gabriel2020go} showed that entity hallucination is the most common type of factual errors made by transformer encoder-decoder models.

We first use Stanza~\citep{qi2020stanza} to perform named entity tagging on both the descriptions and slogans. We limit to the following entity types because they are present in at least 1\% of both the descriptions and slogans based on Table~\ref{tab:entity-stats}: {\fontfamily{qcr}\selectfont GPE}, {\fontfamily{qcr}\selectfont DATE}, {\fontfamily{qcr}\selectfont CARDINAL}, {\fontfamily{qcr}\selectfont LOCATION}, and {\fontfamily{qcr}\selectfont PERSON}. Additionally, we include {\fontfamily{qcr}\selectfont NORP} (nationalities/religious/political group) because a large percentage of entities of this type in the slogan can be found in the corresponding description. We observe that many words are falsely tagged as {\fontfamily{qcr}\selectfont ORGANIZATION}, which is likely because the slogans and descriptions often contain title-case or all-capital texts. Therefore, we exclude {\fontfamily{qcr}\selectfont ORGANIZATION} although it is the most common entity type.

Within each (description, slogan) pair, we maintain a counter for each entity type. We compare each new entity with all previous entities of the same entity type. If it is a substring of a previous entity or vice versa, we assign the new entity to the previous entity's ID. Otherwise, we increment the counter and obtain a new ID. We replace each entity mention with a unique mask token {\fontfamily{qcr}\selectfont[entity\_type]} if it is the first entity of its type or {\fontfamily{qcr}\selectfont[entity\_type\,id]} otherwise. We store a reverse mapping and replace the mask tokens in the generated slogan with the original entity mention. We also apply simple rule-based post-processing, including completing the closing bracket (`{\fontfamily{qcr}\selectfont]}') if it is missing and removing illegal mask tokens and mask tokens not present in the mapping~\footnote{We also remove all the preceding stop words before the removed mask token. In most cases, they are prepositions or articles such as ``\textbf{from the} {\fontfamily{qcr}\selectfont[country]}'' or ``\textbf{in} {\fontfamily{qcr}\selectfont[]}''.}.

During experiments, we observe that when we use the original upper-cased entity type names, the seq2seq model is prone to generating illegal tokens such as {\fontfamily{qcr}\selectfont [gPE]}, {\fontfamily{qcr}\selectfont [GPA]}. Therefore, we map the tag names to a lower-cased word consisting of a single token (as tokenised by the pre-trained tokeniser). The mapping we use is \{{\fontfamily{qcr}\selectfont GPE}:country, {\fontfamily{qcr}\selectfont DATE}:date, {\fontfamily{qcr}\selectfont CARDINAL}:number, {\fontfamily{qcr}\selectfont LOCATION}:location, {\fontfamily{qcr}\selectfont PERSON}:person, {\fontfamily{qcr}\selectfont NORP}:national\}. Table~\ref{tab:entity-masking-example} shows an example of the entity masking process. 

\begin{table}[h!]
\centering
  \caption{An example description and slogan before and after entity masking. Note that the word ``Belgian'' in the slogan is replaced by the same mask token as the same word in the description.}
    \begin{tabular}{lp{10cm}}
    \hline
    \textbf{Description:} & 	
PR-Living \textbf{Belgium} family-owned furniture brand with production facilities in \textbf{Waregem} where it brings the best of \textbf{Belgian}-inspired Design Upholstery \& Furniture pieces to the global consumers.\\
	\textbf{Slogan:} & A \textbf{Belgian} furniture brand \\
    \hdashline
     \textbf{Entities:} & ``Belgium'': {\fontfamily{qcr}\selectfont GPE}, ``Waregem'': {\fontfamily{qcr}\selectfont GPE}, ``Belgian'': {\fontfamily{qcr}\selectfont NORP}\\ 
    \textbf{Masked Description:} & 	
PR-Living {\fontfamily{qcr}\selectfont[country]} family-owned furniture brand with production facilities in {\fontfamily{qcr}\selectfont[country1]} where it brings the best of {\fontfamily{qcr}\selectfont[national]}-inspired
Design Upholstery \& Furniture pieces to the global consumers.\\
    \textbf{Masked Slogan:} & A {\fontfamily{qcr}\selectfont[national]} furniture brand \\ 
    \textbf{Reverse Mapping:} & {\fontfamily{qcr}\selectfont[country]}: ``Belgium'', {\fontfamily{qcr}\selectfont[country1]}: ``Waregem'', {\fontfamily{qcr}\selectfont[national]}: ``Belgian''\\
    \hline
    \end{tabular}
  \label{tab:entity-masking-example}
\end{table}

As shown in Table~\ref{tab:entity-stats}, a sizeable proportion of the entities in the slogans are not present in the description. We discard a (description, slogan) pair from the \textit{training} dataset if any of the entities in the slogan cannot be found in the description. This procedure removes roughly 10\% of the training data but encourages the model to generate entities present in the source description instead of fabricated entities. We do not apply filtering to the validation and test set so that the result is comparable with other models.

\section{Generating Diverse Slogans With Syntactic Control}
\label{sec:syntactic-diversity}

Generating \textbf{diverse} slogans is crucial to avoid ads fatigue and enable personalisation. However, we observe that given one input description, our model tends to generate slogans similar to each other, such as replacing some words or using a slightly different expression. Moreover, the outputs are often simple noun phrases that are not catchy.

To investigate the cause, we perform part-of-speech (POS) tagging on all the slogans in our training dataset. Table~\ref{tab:top-pos-seqs} shows the most frequent POS tag sequences among the slogans~\footnote{We refer readers to {\tt https://www.ling.upenn.edu/courses/Fall\_2003/ling001/penn\_treebank\_pos.html} for the description of each POS tag.}. Only one (\#46) out of the top fifty POS tag sequences is not a noun phrase (VB PRP\$ NN, e.g., Boost Your Business). It motivates us to increase the generated slogans' diversity using syntactic control.

\begin{table}[h!]
  \centering
  \caption{The most frequent ten slogan's POS tag sequences in the training dataset.s}
    \begin{tabular}{lcl}
    \hline
   	\textbf{POS Tag Sequence} & \textbf{Frequency} & \textbf{Example} \\\hline
	NNP NNP NNP & 12,892 & Emergency Lighting Equipment \\\hdashline
	NNP NNP NNP NNP & 5,982 & Personal Injury Lawyers Melbourne \\\hdashline
	NNP NNP NNPS & 4,217 & Rugged Computing Solutions \\\hdashline
	JJ NN NN & 3,109 & Flexible Office Space \\\hdashline
	NNP NNP NN & 2789 & Bluetooth Access Control \\\hdashline
	NNP NNP NNS & 2,632 & Manchester Law Firms \\\hdashline
	NNP NNP NNP CC NNP NNP & 2,190 & Local Programmatic Advertising \& DSP Platform \\\hdashline
	NNP NNP CC NNP NNP NNP & 2,157 & Retro Candy \& Soda Pop Store \\\hdashline
	NN NN NN & 2,144 & Footwear design consultancy \\\hdashline
	NNP NNP NNP NNP NNP & 1,662 & Commercial Construction Company New England \\
    \hline
    \end{tabular}
  \label{tab:top-pos-seqs}
\end{table}

Inspired by CTRL~\citep{keskarCTRL2019}, we modify the generation from $P(slogan \vert description)$ to $P(slogan \vert description, ctrl)$ by conditioning on an additional syntactic control code. To keep the cardinality small, we use the coarse-grained POS tag~\footnote{Corresponding to the first two characters of the POS tag, thus ignoring the difference between proper vs common noun, plural vs singular, different verb tenses and the degree of adjectives and adverbs.} of the first word in the slogan as the control code. Additionally, we merge adjectives and adverbs and merge all the POS tags that are not among the most frequent five tags. Table~\ref{tab:ctrl-codes} shows the full list of control codes.

\begin{table}[h!]
  \centering
  \caption{Full list of syntactic control codes.}
    \begin{tabular}{lcp{8cm}}
    \hline
   	\textbf{Code} & \textbf{Frequency} & \textbf{Meaning} \\\hline
	NN & 208,061 & All types of nouns \\\hdashline
	JJ & 44,926 & All types of adjectives and adverbs \\\hdashline
	VB & 37,331 & Verbs of any form or tense \\\hdashline
	DT & 17,645 & Determiners \\\hdashline
	PR & 8,484 & Personal or possessive pronouns \\\hdashline
	OTHER & 7,644 & Any other tags not included above, such as numbers, prepositions, and question words \\
    \hline
    \end{tabular}
  \label{tab:ctrl-codes}
\end{table}

While we can use the fine-grained POS tags or even the tag sequences as the control code, they have a long-tail distribution, and many values have only a handful of examples, which are too few for the model to learn from. \citet{munigala2018persuaide} applied a similar idea as ours to generate persuasive text starting with a verb. However, they apply rules to restrict a generic language model to start with a verb. We apply conditional training to learn the characteristics of slogans starting with words belonging to various POS tags.

We prepend the control code to the input sequence with a special {\fontfamily{qcr}\selectfont</s>} token separating the control code and the input sequence. We use the control code derived from the target sequence during training while we randomly sample control codes during inference to generate syntactically diverse slogans. Our method differs from \citet{keskarCTRL2019} in two slight ways: 1) CTRL uses an autoregressive transformer similar to GPT-2~\citep{radford2019language} while we use an encoder-decoder transformer with a bi-directional encoder. 2) The control codes were used during pre-training in CTRL while we prepend the control code only during fine-tuning for slogan generation. 

\section{Experiments}
\label{sec:experiments}

We conduct a comprehensive evaluation of our proposed method. In Section~\ref{subsec:exp:quantitative}, we conduct a quantitative evaluation and compare our proposed methods with other rule-based and encoder-decoder baselines in terms of ROUGE -1/-2/-L F$_1$ scores. We report the performance of a larger model in Section~\ref{subsec:larger-model}. We specifically study the truthfulness and diversity of the generated slogans in Section~\ref{subsec:factual-eval} and Section~\ref{subsec:diversity-eval}. Finally, we conduct a fine-grained human evaluation in Section~\ref{subsec:human-evaluation} to further validate the quality of the slogans generated by our model. 

We use the DistilBART and BART$_{LARGE}$ implementation in the Hugging Face library~\citep{wolf2019huggingface} with a training batch size of 64 for DistilBART and 32 for BART$_{LARGE}$. We use a cosine decay learning rate with warm-up~\citep{he2019bag} and a maximum learning rate of 1e-4. The learning rate is chosen with Fastai's learning rate finder~\citep{howard2020fastai}.

We train all BART models for three epochs. Based on our observation, the models converge within around 2-3 epochs. We use greedy decoding unless otherwise mentioned. We also add a repetition penalty $\theta=1.2$ following \cite{keskarCTRL2019}. 

\subsection{Quantitative Evaluation}
\label{subsec:exp:quantitative}

We leave the diversity evaluation to Section~\ref{subsec:diversity-eval} because we have only a single reference slogan for each input description in our dataset, which will penalise systems generating diverse slogans. We compare our proposed method with the following five baselines:

\begin{itemize}
\setlength\itemsep{1em}
  \item \textit{first sentence}: predicting the first sentence from the description as the slogan, which is simple but surprisingly competitive for document summarization~\citep{katragadda2009sentence}. We use the sentence splitter in the Spacy library~\footnote{\tt https://spacy.io/} to extract the first sentence.
  \item \textit{first-k words}: predicting the first-$k$ words from the description as the slogan. We choose $k$ that yields the highest ROUGE-1 F$_1$ score on the validation dataset. We add this baseline because the first sentence of the description is usually much longer than a typical slogan.
  \item \textit{Skeleton-Based}~\citep{tomavsic2014implementation}: a skeleton-based slogan generation system using genetic algorithms and various heuristic-based scoring functions. We sample a random compatible slogan skeleton from the training dataset and realise the slogan with keywords extracted from the company description. We follow \citet{tomavsic2014implementation}'s implementation closely. However, we omit the database of frequent grammatical relations and the bigram function derived from Corpus of Contemporary American English because the resources are not available.
  \item \textit{Encoder-Decoder}~\citep{bahdanau2015neural}: a strong and versatile GRU encoder-decoder baseline. We use identical hyper-parameters as \cite{misawa2020distinctive} and remove the reconstruction loss and copying mechanism to make the models directly comparable. Specifically, the model has a single hidden layer for both the bi-directional encoder and the auto-regressive decoder. We apply a dropout of 0.5 between layers. The embedding and hidden dimensions are 200 and 512 separately, and the vocabulary contains 30K most frequent words. The embedding matrix is randomly initialised and trained jointly with the model. We use Adam optimiser with a learning rate of 1e-3 and train for 10 epochs (The encoder-decoder models take more epochs to converge than the transformer models, likely because the models are randomly initialised).
  \item \textit{Pointer-Generator}~\citep{see2017get}: encoder-decoder model with copying mechanism to handle unknown words. Equivalent to \cite{misawa2020distinctive} with the reconstruction loss removed.
  \item \cite{misawa2020distinctive}: a GRU encoder-decoder model for slogan generation with additional reconstruction loss to generate distinct slogans and copying mechanism to handle unknown words.
\end{itemize}

Table~\ref{tab:performance} presents the ROUGE -1/-2/-L scores of various models on both the validation and the manually-curated test dataset.
 
\begin{table}[h!]
  \centering
  \caption{The ROUGE F$_1$ scores for various models on the validation dataset and the test dataset. DistilBART denotes the base model introduced in Section~\ref{sec:model}. ``+delex'' and ``+ent'' means adding company name delexicalisation (Section~\ref{subsec:method:delexicalisation}) and entity masking (Section~\ref{subsec:factual-entity}).}
    \begin{tabular}{lrrrrrr}
    \hline
   	& \multicolumn{3}{c}{Valid Dataset} & \multicolumn{3}{c}{Test Dataset}  \\
   	& R1 & R2 & RL & R1 & R2 & RL \\\hline
   	First sentence & 26.12 & 13.03 & 23.88 & 25.50 & 12.73 & 23.47 \\ 
   	First-$k$ words ($k=11$) & 27.50 & 13.72 & 25.33 & 25.76 & 12.68 & 24.02 \\
   	Skeleton-based & 16.09 & 1.62 & 14.01 & 16.61 & 1.79 & 14.72 \\
	Encoder-Decoder & 24.85 & 9.38 & 24.01 & 23.91 & 9.31 & 23.28 \\
	Pointer-Generator & 26.42 & 10.15 & 25.63 & 26.24 & 10.67 & 25.65 \\
	\cite{misawa2020distinctive} & 24.14 & 9.19 & 23.37 & 26.01 & 10.00 & 25.39 \\\hdashline
	DistilBART & 36.74 & 18.87 & 33.97 & 34.95 & 17.38 & 32.47 \\
	DistilBART+delex & 37.37 & 19.51 & 34.69 & 35.06 & 17.79 & 32.52 \\
	DistilBART+delex+ent & \textbf{37.76} & \textbf{19.69} & \textbf{35.17} & \textbf{35.58} & \textbf{18.47} & \textbf{33.32} \\
    \hline
    \end{tabular}
  \label{tab:performance}
\end{table}

\begin{figure}
  \centering
  \includegraphics[width=.6\textwidth]{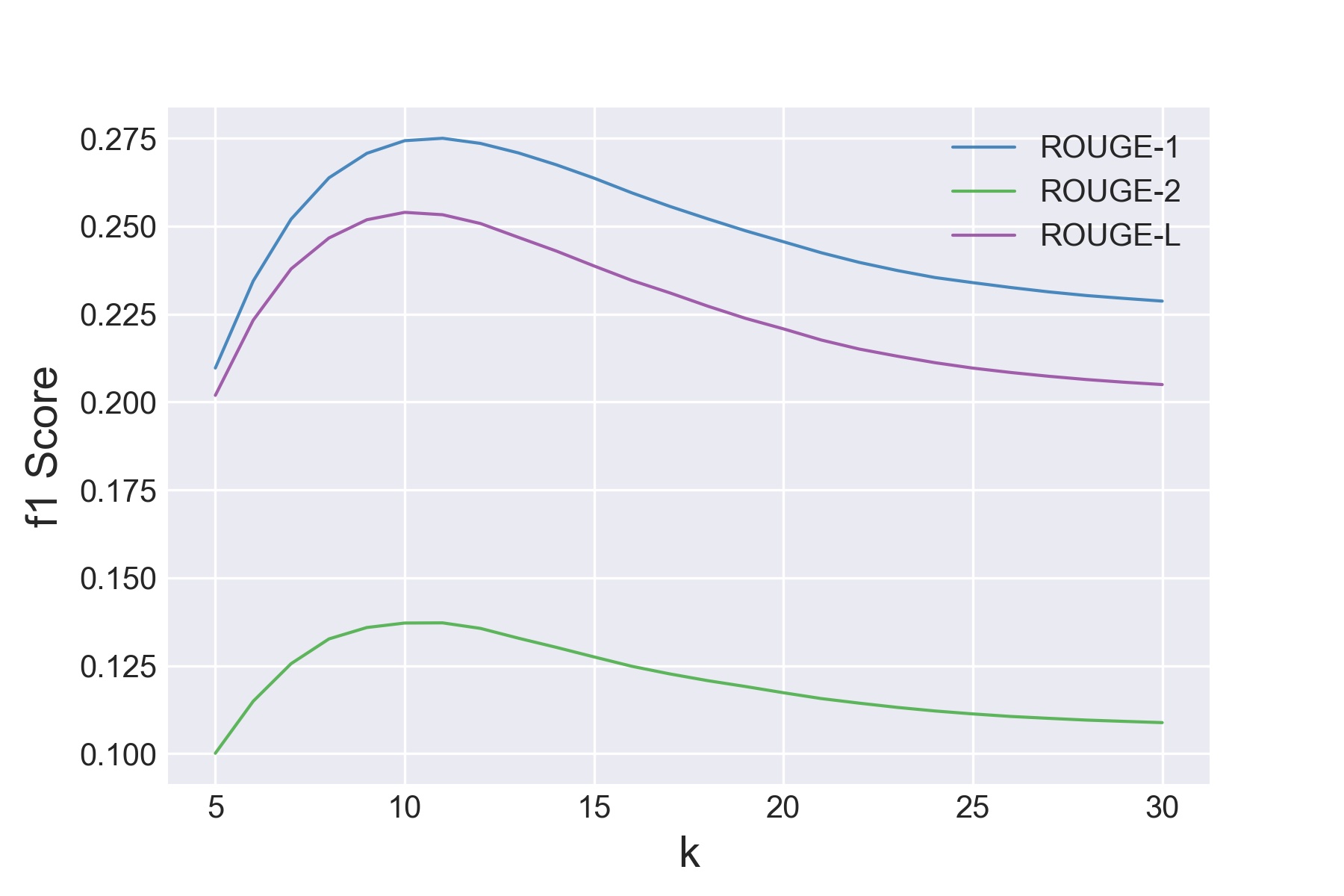}
\caption{The ROUGE -1/-2/-L scores of the first-$k$ word baseline by varying $k$.}
\label{fig:first-k-baseline}
\end{figure}

The first-$k$ words baseline achieved a reasonable performance, showing a certain degree of overlap between slogans and descriptions. Figure~\ref{fig:first-k-baseline} shows how the first-$k$ words baseline's ROUGE F$_1$ scores change by varying $k$. It is obvious that not the larger $k$, the better. The best ROUGE scores are achieved when $k$ is in the range (9, 12). The first-$k$ words baseline also achieved higher ROUGE scores than the first sentence baseline, although it may output an incomplete phrase due to the truncation. 

The skeleton-based method had the worst performance among all baselines. While it often copies important keywords from the description, it is prone to generating ungrammatical or nonsensical output because it relies on POS sequence and dependency parse skeletons and ignores the context.

Comparing the three GRU encoder-decoder baselines, it is clear that the copying mechanism in Pointer-Generator improved the ROUGE scores consistently. However, the reconstruction loss introduced in \citet{misawa2020distinctive} seems to reduce performance. We hypothesise that the slogan is much shorter than the input description. Therefore, reconstructing the description from the slogan may force the model to attend to unimportant input words. Overall, the Pointer-Generator baseline's performance is on par with the first-$k$ words baseline but pales when comparing with any transformer-based model. 

Both delexicalisation and entity masking further improved DistilBART's performance. The final model achieved a ROUGE -1/-2/-L score of 35.58/18.47/33.32 on the curated test set, outperforming the best GRU encoder-decoder model by almost 10\% in ROUGE score. Table~\ref{tab:compare_systems} provides a more intuitive overview of various models' behaviour by showing the generated slogans from randomly sampled company descriptions.

\begin{table}[h!]
  \centering
  \caption{Sample generated slogans by various systems. ``Gold'' is the original slogan of the company. The DistilBART model uses both delexicalisation and entity masking.}
    \begin{tabular}{lp{10cm}}
    \hline
   	\textbf{Gold:} & Fast, Fresh \& Tasty Mexican Food \\
   	\textbf{First-k words:} & We may not be the only burrito in town, but we've \\
   	\textbf{Skeleton-based:} & Bar to Your Burrito in Town \\
    \textbf{Pointer-Generator:} & The World 's First Class Action Club\\
    \textbf{DistilBART:} & The Best Burrito in Town \\\hdashline
    \textbf{Gold:} & A better UK energy supplier \\
   	\textbf{First-k words:} & Welcome to Powershop, a better gas and energy supplier. We offer \\
   	\textbf{Skeleton-based:} & Your Gas in Competitive Electricity, Energy Deals, Offer and Website \\
    \textbf{Pointer-Generator:} & Gas and Electricity Supplier\\
    \textbf{DistilBART:} & Gas and Energy Supplier \\\hdashline
    \textbf{Gold:} & Saigon Food Tours and City Tours Led by Women \\
   	\textbf{First-k words:} & Top-ranked Ho Chi Minh City food tours from the first company \\
   	\textbf{Skeleton-based:} & Food Company \& Culture Tours Female \\
    \textbf{Pointer-Generator:} & Food Tours \& Food Tours\\
    \textbf{DistilBART:} & Food Tours in Vietnam \\\hdashline
    \textbf{Gold:} & Top Engineering Colleges In Tamilnadu \\
   	\textbf{First-k words:} & top Engineering colleges in Tamil Nadu based on 2020 ranking. Get \\
   	\textbf{Skeleton-based:} & We Info Colleges! \\
    \textbf{Pointer-Generator:} & Engineering College in Dehradun Uttarakhand\\
    \textbf{DistilBART:} & Top Engineering Colleges in Tamil Nadu \\\hline
    \end{tabular}
  \label{tab:compare_systems}
\end{table}

We can observe that while the first-$k$ words baseline sometimes has substantial word overlap with the original slogan, its style is often different from slogans. Pointer-Generator and DistilBART sometimes generate similar slogans. However, Pointer-Generator is more prone to generating repetitions, as in the third example. It also hallucinates much more. In the first example, the company is a Mexican restaurant. The slogan generated by Pointer-Generator is fluent but completely irrelevant. In the last example, it hallucinated the location of the school while DistilBART preserved the correct information.

\subsection{Larger Model Results}
\label{subsec:larger-model}

Following \citet{lewis-etal-2020-bart} and \citet{zhang2020pegasus}, we report the performance of a larger model, BART$_{LARGE}$~\footnote{\tt https://huggingface.co/facebook/bart-large-cnn}. Compared to DistilBART, BART$_{LARGE}$ has both more layers ($L$: 6 $\rightarrow$ 12) and a larger hidden size ($H$: 768 $\rightarrow$ 1024). We follow the exact training procedure as DistilBART. Table~\ref{tab:performance-large-scale} compares the performance of DistilBART and BART$_{LARGE}$.

\begin{table}[h!]
  \centering
  \caption{The ROUGE F$_1$ scores by scaling up the model size. Both models use delexicalisation and entity masking.}
    \begin{tabular}{lcccccccc}
    \hline
   	& & &\multicolumn{3}{c}{Valid Dataset} & \multicolumn{3}{c}{Test Dataset}  \\
   	& params & epoch time & R1 & R2 & RL & R1 & R2 & RL \\\hline
	DistilBART& 230M & 45 mins &\textbf{37.76} & \textbf{19.69} & \textbf{35.17} & \textbf{35.58} & \textbf{18.47} & \textbf{33.32} \\\hdashline
	BART$_{LARGE}$ & 406M & 86 mins & 35.36 & 17.71 & 32.95 & 33.43 & 16.51 & 31.34 \\\hline
    \end{tabular}
  \label{tab:performance-large-scale}
\end{table}

We were surprised to observe that BART$_{LARGE}$ underperformed the smaller DistilBART model by roughly 2\% ROUGE score. During training, BART$_{LARGE}$ had a lower training loss than DistilBART but the validation loss plateaued to roughly the same value, suggesting that the large model might be more prone to overfitting the training data. We did not conduct extensive hyperparameter tuning and used the same learning rate as DistilBART. Although we cannot conclude that DistilBART is better suited for this task, it seems that using a larger model does not always improve performance.

\subsection{Truthfulness Evaluation}
\label{subsec:factual-eval}

In this Section, we employ automatic truthfulness evaluation metrics to validate that the methods proposed in Section~\ref{sec:improve-factual} indeed improved the truthfulness. As briefed in Section~\ref{subsec:literature-factual}, there are mainly three categories of automatic truthfulness evaluation metrics, namely entailment, information extraction, and QA. We focus on entailment-based metrics because 1) they yield the highest correlation with human judgement based on a recent benchmark~\citep{pagnoni2021understanding}, 2) the slogans are often very short and sometimes do not contain a predicate, making it impossible to automatically generate questions for a QA-based approach and extract (subject, verb, object) tuples for an information extraction-based approach. 

The first model we use is an entailment classifier fine-tuned on the Multi-Genre NLI (MNLI) dataset~\citep{williams2018broad} following \citet{maynez2020faithfulness}. However, we use a fine-tuned RoBERTa$_{LARGE}$ checkpoint~\footnote{\tt https://huggingface.co/roberta-large-mnli}~\citep{liu2019roberta} instead of BERT$_{LARGE}$~\citep{devlin2019bert} since it achieved higher accuracy on the MNLI dataset (90.2 vs. 86.6). We calculate the entailment probability between the input description and the generated slogan to measure truthfulness.

The second model we use is a pre-trained FactCC~\citep{kryscinski2020evaluating} classifier, which predicts whether a generated summary is consistent with the source document. It was trained on a large set of synthesised examples by adding noise into reference summaries using manually defined rules such as entity or pronoun swap. FactCC is the best-performing metric in ~\citet{pagnoni2021understanding}'s benchmark. It was also used in several subsequent works as the automatic truthfulness evaluation metric~\citep{dong2020multi,cao2020factual}. We use the predicted probability for the category ``consistent'' to measure truthfulness.

Table~\ref{tab:performance-factuality} presents the mean entailment and FactCC scores for both the validation and the test dataset. Both metrics suggest that our proposed method yields more truthful slogans w.r.t the input descriptions than a DistilBART baseline with strong statistical significance.

\begin{table}[h!]
  \centering
  \caption{The truthfulness scores of the baseline distillBART model and our proposed method (the numbers are in per cent). The p-value of a two-sided paired t-test is shown in the bracket.}
    \begin{tabular}{lllll}
    \hline
   	&\multicolumn{2}{c}{Valid Dataset} & \multicolumn{2}{c}{Test Dataset}  \\
    & Entailment & FactCC & Entailment & FactCC \\\hline
	DistilBART & 75.89 & 70.23 & 70.87 & 71.21 \\\hdashline
	DistilBART+delex+ent & \textbf{83.25} (2.3e-63) & \textbf{73.09} (5.1e-6) & \textbf{81.61} (1.0e-21) & \textbf{75.71} (8.4e-4) \\\hline
    \end{tabular}
  \label{tab:performance-factuality}
\end{table}

Compared to the result in Section~\ref{subsec:exp:quantitative}, there is a larger gap between our proposed method and the baseline DistilBART model. It is likely because n-gram overlap metrics like ROUGE are not very sensitive to local factual errors. E.g., suppose the reference sequence is ``Digital Marketing Firm in New \textit{Zealand}'', and the predicted sequence is ``Digital Marking Firm in New \textit{Columbia}'', it will receive a high ROUGE-1/-2/-L score of 83.3/80.0/83.3. However, entailment and factuality models will identify such factual inconsistencies and assign a very low score.

\subsection{Diversity Evaluation}
\label{subsec:diversity-eval}

In Section~\ref{sec:syntactic-diversity}, we proposed a method to generate syntactically diverse slogans using control codes. First, we want to evaluate whether the control codes are effective in the generation. We calculate the \textit{ctrl accuracy}, which measures how often the first word in the generated slogan agrees with the specified POS tag.

We apply each of the six control codes to each input in the test set and generate various slogans using greedy decoding. We then apply POS tagging on the generated slogans and extract the coarse-grained POS tag of the first word in the same way as in Section~\ref{sec:syntactic-diversity}. We count it as successful if the coarse-grained POS tag matches the specified control code. Table~\ref{tab:diversity-eval} presents the ctrl accuracy for each of the control codes.

\begin{table}[h!]
  \centering
  \caption{The syntactic control accuracy, diversity and abstractiveness scores of various methods. The best score for each column is highlighted in bold. All models use neither delexicalisation nor entity masking to decouple the impact of different techniques.}
    \begin{tabular}{lrrrrrrrr}
    \hline
   	&\multicolumn{6}{c}{Ctrl Accuracy} & Diversity & Abstractive \\
    & NN & JJ & VB & DT & PR & OTHER &  &  \\\hline
	W/O Upsampling & \textbf{92.56} & 37.12 & \textbf{61.47} & \textbf{93.96} & \textbf{97.28} & \textbf{90.64} & \textbf{46.69} & \textbf{45.04} \\\hdashline
	Upsampling  & 91.14 & \textbf{42.35} & 48.69 & 71.83 & 96.88 & 55.53 & 44.81 & 43.85 \\\hdashline
	Nucleus Sampling & 70.32 & 11.27 & 7.85 & 5.23 & 0.80 & 1.31 & 27.97 & 27.01 \\\hline
    \end{tabular}
  \label{tab:diversity-eval}
\end{table}

The control code distribution in our training dataset is very skewed, as shown in Table~\ref{tab:ctrl-codes}. The most frequent code (NN) contains more than 27 times more data than the least frequent code (OTHER). Therefore, we conducted another experiment by randomly upsampling examples with codes other than NN to 100k. We then trained for one epoch instead of three epochs to keep the total training steps roughly equal. We show the result in the second row of Table~\ref{tab:diversity-eval}. 

Besides, we compare with the nucleus sampling~\citep{holtzman2019curious} baseline. We use top-$p=0.95$ following \citet{holtzman2019curious} because it is scaled to match the human perplexity~\footnote{We use the default temperature of 1.0 and disable the top-$k$ filter.}. We generate an equal number of slogans (six) as our method, and the result is presented in the third row of Table~\ref{tab:diversity-eval}. We note that since nucleus sampling does not condition on the control code, the ctrl accuracies are not meant to be compared directly with our method but serve as a random baseline without conditional training.

We calculate the diversity as follows: for each set of generated slogans from the same input, we count the total number of tokens and unique tokens. We use Spacy's word tokenisation instead of the subword tokenisation. Besides, we lowercase all words, so merely changing the case will not be counted towards diversity. The diversity score for each set is the total number of unique tokens divided by the total number of tokens. We average the diversity scores over the whole test set to produce the final diversity score. We note that a diversity score of close to 100\% is unrealistic because important keywords and stop words will and should occur in various slogans. However, a diversity score of close to 1/6 (16.67\%) indicates that the model generates almost identical slogans and has very little diversity.

The result shows that our method achieved close to perfect ctrl accuracy except for the control code JJ and VB. Although some control codes like PR and OTHER have much fewer examples, they also have fewer possible values and are easier to learn than adjectives and verbs (e.g., there are a limited number of pronouns). The strong syntactic control accuracy validated recent studies' finding that pretrained language models capture linguistic features internally~\citep{tenney2019bert,rogers2020primer}.

Upsampling seems to help with neither the ctrl accuracy nor the diversity. Compared with our method, nucleus sampling has much lower diversity. Although it performs sampling among the top-$p$ vocabulary, it will almost always sample the same words when the distribution is peaked. Increasing the temperature to above 1.0 can potentially increase the diversity, but it will harm the generation quality and consistency~\citep{holtzman2019curious}.

In addition, we calculate the abstractiveness as the number of generated slogan tokens that are not present in the input description divided by the number of generated slogan tokens, averaging over all candidates and examples in the test set. We can see that as a by-product of optimising towards diversity, our model is also much more abstractive.

Finally, we invite an annotator to manually assess the quality of the generated slogans~\footnote{The annotator is an NLP researcher who is proficient in English and was not involved in the development of this work.}. We randomly sample 50 companies from the test set and obtain the six generated slogans from both our proposed method and nucleus sampling, thus obtaining 300 slogan pairs. We then ask the annotator to indicate which slogan is better (with the ``can't decide'' option). We randomised the order of the slogans to eliminate positional bias. We present the annotation UI in Appendix~\ref{apx:annotation-ui} and the annotation result in Table~\ref{tab:diversity-eval-result}.

\begin{table}[h!]
  \centering
  \caption{Pair-wise evaluation result of each control code vs the nucleus sampling baseline. The p-value is calculated using two-sided Wilcoxon signed-rank test. ``Better'' means the annotator indicates that the slogans generated by our method is better than nucleus sampling.}
    \begin{tabular}{rrrrr}
    \hline
   	\textbf{Code} & \textbf{Better} & \textbf{Can't Decide}  & \textbf{Worse} & \textbf{p-value} \\\hline
	NN & 28 & 3 & 19 & 0.189\\
	JJ & 32 & 3 & 15 & 0.013 \\
	VB & 36 & 0 & 14 & 1.86e-03 \\
	DT & 41 & 1 & 8 & 2.43e-06\\
	PR & 37 & 0 & 13 & 6.89e-04\\
	OTHER & 39 & 0 &  11 & 7.50e-05\\\hdashline
	Overall & 213 & 7 & 80 & 7.85e-15\\\hline
    \end{tabular}
  \label{tab:diversity-eval-result}
\end{table}

All control codes except ``NN'' yielded significantly better slogans than the nucleus sampling baseline with $p=0.05$. It is expected because ``NN'' is most common in the dataset, and using the control code ``NN'' will yield similar output as greedy decoding or nucleus sampling. While \citet{munigala2018persuaide} claimed that sentences starting with a verb is more persuasive, sentences starting with other POS tags may also have desirable characteristics for slogans. E.g., Starting with an adjective makes it more vivid; starting with a determiner makes it more assertive; starting with a pronoun makes it more personal. Surprisingly, the annotator also rated slogans generated with the control code ``OTHER'' highly despite it groups many long-tail POS tags. The ``OTHER'' control code often generates slogans starting with a question word, an ordinal number (e.g., ``\#1'') or the preposition ``for'' (e.g., ``For All Your Pain Relief Needs'').

To give the reader a better sense of the system's behavior, we present samples the system generated with different control codes in Table~\ref{tab:diverse-examples}. We can see that the first word in the slogan may not always match the POS tag specified by the control code. However, the generated slogans are diverse in both syntactic structure and content.

\begin{table}[h!]
  \centering
  \caption{Sample generated slogans with different control codes (no cherry-picking).}
    \begin{tabular}{rp{11cm}}
    \hline
   	\textbf{Desc:} & Helping eCommerce business growing their sales \& revenues. Specialist in product feeds, shopping ads, conversion optimization, SEO and website personalization. \\\hdashline
   	\textbf{NN:} & eCommerce Business Growth \& Revenue Optimization Experts\\
   	\textbf{JJ:} & Ecommerce Marketing Agency in London \& Essex \\
   	\textbf{VB:} & Helping eCommerce Business Grow Their Sales \& Revenues \\
   	\textbf{DT:} & The eCommerce Experts \\
   	\textbf{PR:} & Your eCommerce Partner for Growth \& Success! \\
   	\textbf{OTHER:} &  How to Grow Your Business with eCommerce \\\hdashline
	\textbf{Desc:}  & We are experts in: Web / Mobile / Desktop apps Development. Innovative technologies. \\\hdashline
   	\textbf{NN:}  & Web and Mobile App Development Company in India\\
   	\textbf{JJ:}  & Innovative Technologies. Web and Mobile Apps Development Company \\
   	\textbf{VB:}  & Leading Mobile App Development Company in India \\
   	\textbf{DT:}  & Achieving Digital Transformation in the Cloud with Mobile Apps Development \\
   	\textbf{PR:}  & We are experts in mobile apps development \\
   	\textbf{OTHER:}  & Where technology meets creativity \\
    \hline
    \end{tabular}
  \label{tab:diverse-examples}
\end{table}

Besides generating more diverse and higher quality slogans, another principal advantage of our approach over nucleus sampling is that we have more control over the syntactic structure of the generated slogan instead of relying purely on randomness. 

\subsection{Human Evaluation}
\label{subsec:human-evaluation}

Based on the evaluation we conducted in previous sections, we include all the methods we introduced in Section~\ref{sec:improve-factual} and~\ref{sec:syntactic-diversity} in our final model, namely, company name delexicalisation, entity masking and conditional training based on the POS tag of the first slogan token. We incorporate an additional control code ``ENT'' to cover the cases where a reference slogan starts with an entity mask token. Based on the result in Section~\ref{subsec:diversity-eval}, we randomly sample a control code from the set \{JJ, VB, DT, PR, OTHER\} during inference time. Finally, we replace the entity mask tokens in the slogan (if any) using the reverse dictionary induced from the input description to produce the final slogan as described in Section~\ref{subsec:factual-entity}.

We randomly sampled 50 companies from the test set (different from the sample in Section~\ref{subsec:diversity-eval}) and obtained the predicted slogans from our model, along with four other baselines: first sentence, skeleton-based, Pointer-Generator, and DistilBART. Therefore, we have in total 250 slogans to evaluate. We invited two human annotators to score the slogans independently based on three fine-grained aspects: coherence, well-formedness, and catchiness. They assign scores on a scale of 1-3 (poor, acceptable, good) for each aspect.

We display the input description along with the slogan so that the annotators can assess whether the slogan is coherent with the description.
We also randomise the slogans' order to remove positional bias. The annotation guideline is shown in Appendix~\ref{apx:annotation-guideline} and the annotation UI is presented in Appendix~\ref{apx:annotation-ui}. 

\begin{figure}
  \centering
  \includegraphics[width=.5\textwidth]{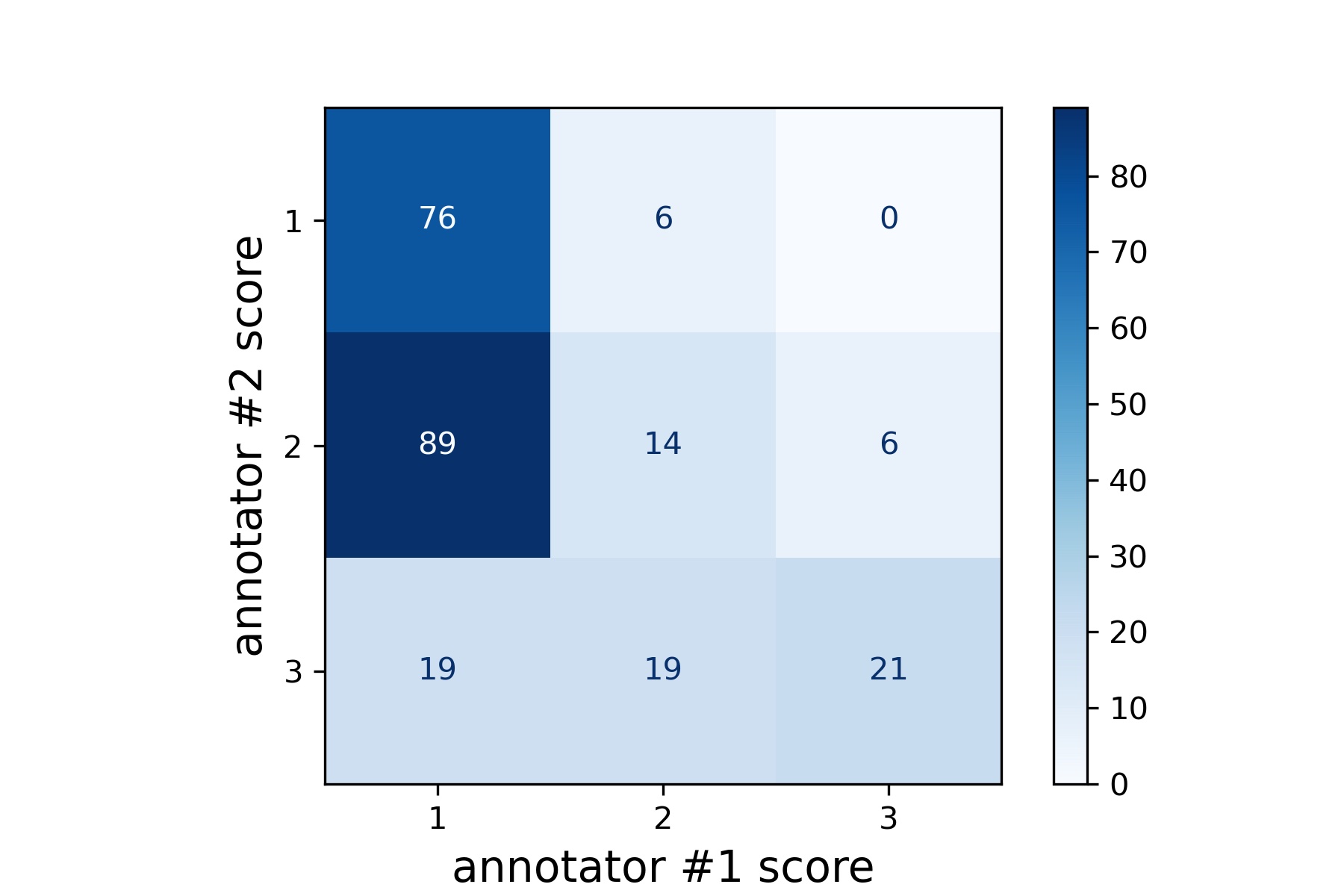}
\caption{Confusion matrix of the catchiness scores assigned by the two annotators.}
\label{fig:confusion_catchiness}
\end{figure}

We measure the inter-annotator agreement using Cohen's kappa coefficient~\citep{cohen1960coefficient}. The $\kappa$ value for coherence, well-formedness, and catchiness are 0.493 (moderate), 0.595 (moderate), and 0.164 (slight) separately. The ``catchiness'' aspect has a low $\kappa$ value because it is much more subjective. While the annotators generally agree on an unattractive slogan, their standards for catchiness tend to differ. It can be illustrated in Figure~\ref{fig:confusion_catchiness} where the agreement is high when the assigned score is 1 (poor). However, there are many examples where annotator 1 assigned score 1 (poor) and annotator 2 assigned score 2 (acceptable). There are only 19 slogans (7.6\%) where the annotators assigned opposite labels. Therefore, we believe the low agreement is mainly due to individual differences rather than annotation noise.

We average the scores assigned by the two annotators and present the result in Table~\ref{tab:human-eval}.

\begin{table}[h!]
  \centering
  \caption{Human evaluation on three aspects: coherence, well-formedness, and Catchiness. We average the scores assigned by the two annotators. The best score for each aspect is highlighted in bold (we exclude the first sentence baseline for the ``coherent'' aspect because it is ``coherent'' by definition). ** indicates statistical significance using a double-sided paired t-test  with p-value=0.005 comparing with our proposed method.}
    \begin{tabular}{llll}
    \hline
   	\textbf{System} & \textbf{Coherent} & \textbf{Well-formed}  & \textbf{Catchy} \\\hline
	First sentence & 3.00$^{**}$ & 1.49$^{**}$ & 1.19$^{**}$ \\
	Skeleton-based & 2.31$^{**}$ & 1.36$^{**}$ & 1.41$^{**}$  \\
	Pointer-Generator & 2.63$^{**}$& 2.07$^{**}$ & 1.51$^{**}$ \\
	DistilBART & 2.89 & \textbf{2.81} & 1.87$^{**}$ \\\hdashline
	Ours & \textbf{2.91} & 2.79 & \textbf{2.22} \\\hline
    \end{tabular}
  \label{tab:human-eval}
\end{table}

The first sentence baseline received low well-formedness and catchiness scores. As we mentioned earlier, the first sentence of the description is often much longer than a typical slogan, failing to satisfy the conciseness property of slogans. \citet{lucas1934optimum} observed that longer slogans are also less memorable and attractive, which is validated by the low catchiness score. 

The skeleton-based approach improved the catchiness over the first sentence baseline by a reasonable margin. However, it received the lowest well-formedness score due to the limitations of skeletons, causing it to generate nongrammatical or nonsensical slogans occasionally. Moreover, it has a much lower coherence score than either LSTM or Transformer seq2seq models, which is our primary motivation to apply the seq2seq framework instead of relying on random slogan skeletons. 

The Pointer-Generator baseline outperformed the previous two baselines across all aspects. On the one hand, it demonstrates the capability of modern deep learning models. On the other hand, it surfaces the limitations of word overlap-based evaluation metrics. Based on the ROUGE scores reported in Section~\ref{subsec:exp:quantitative} alone, we could not conclude the superiority of the Pointer-Generator model over the first sentence or first-$k$ words baseline.

The DistilBART model improved further over the Pointer-Generator baseline, especially in the well-formedness aspect. It is likely due to the extensive pre-training and its ability to generate grammatical and realistic text.

Our proposed method received similar coherence and well-formedness scores as DistilBART. However, it outperformed all other methods in catchiness by a large margin. Although the improvement of coherence is not statistically significant, it does not necessarily mean the delexicalisation and entity masking techniques are not helpful. As we discussed in Section~\ref{subsec:diversity-eval}, our method generates substantially more diverse slogans, and the generation is much more abstractive than the DistilBART baseline. Previous work highlighted the trade-off between abstractiveness and truthfulness~\citep{durmus2020feqa}. By combining the approaches to improve truthfulness and diversity, our proposed method generates more catchy and diverse slogans without sacrificing truthfulness or well-formedness.

\section{Ethical Considerations}
\label{sec:ethical}

All three annotators employed in this study are full-time researchers at Knorex. We explained to them the purpose of this study and obtained their consent. They conducted the annotation during working hours and are paid their regular wage.

Marketing automation is a strong trend in the digital advertising industry. AI-based copywriting is a challenging and crucial component in this process. We take generating counter-factual advertising messages seriously as it might damage the advertiser's brand image and harm the prospects. The model proposed in this work generates better-quality and more truthful slogans than various baselines. However, we cannot yet conclude that the generated slogans are 100\% truthful, just like most recently proposed language generation models. This work is being integrated into a commercial digital advertising platform~\footnote{\tt knorex.com}. In the initial version, advertisers are required to review and approve the slogans generated by the system. They can also make modifications as necessary before the ads go live.

\section{Conclusion}
\label{sec:conclusion}
In this work, we model slogan generation using a sequence-to-sequence transformer model with the company's description as input. It ensures coherence between the generated slogan and the company's marketing communication. In addition, we applied company name delexicalisation and entity masking to improve the generated slogans' truthfulness. We also introduced a simple conditional training method to generate more diverse slogans. Our model achieved a ROUGE -1/-2/-L F$_1$ score of 35.58/18.47/33.32 on a manually-curated slogan dataset. Comprehensive evaluations demonstrated that our proposed method generates more truthful and diverse slogans. A Human evaluation further validated that the slogans generated by our system are significantly catchier than various baselines.

As ongoing work, we are exploring other controllable aspects, such as the style~\citep{jin2020hooks} and the sentence parse~\citep{sun2021aesop}. Besides, we are also working on extending our method to generating longer texts~\citep{hua2021dyploc} which can be used as the body text in advertising.

\section*{Acknowledgement}
Yiping is supported by the scholarship from ``The 100$^{th}$ Anniversary Chulalongkorn University Fund for Doctoral Scholarship'' and also ``The 90$^{th}$ Anniversary Chulalongkorn University Fund (Ratchadaphiseksomphot Endowment Fund)''. We would like to thank our colleagues Vishakha Kadam, Khang Nguyen and Hy Dang for conducting the manual evaluation on the slogans. We would like to thank the anonymous reviewers for their careful reading of the manuscript and constructive criticism. 

\bibliographystyle{nlelike}
\bibliography{nle}

\clearpage

\appendix

\section{Details of Data Cleaning}
\label{apx:data-cleaning}

We perform the following steps in sequence to obtain clean (description, slogan) pairs.

\begin{enumerate}
\setlength\itemsep{0.5em}
  \item Delexicalise the company name in both the description and the HTML page title.
  \item Remove all non-alphanumeric characters at the beginning and the end of the HTML page title.
  \item Filter by blocked keywords/phrases. Sometimes the crawling is blocked by a firewall, and the returned title is ``Page could not be loaded'' or ``Access to this page is denied''. We did a manual analysis of a large set of HTML page titles and came up with a list of 50 such blocked keywords/phrases.
  \item Remove prefix or suffix phrases indicating the structure in the website, such as ``Homepage - '', ``$\vert$ Welcome page'', ``About us''. 
  \item Split the HTML page title with special characters~\footnote{We use one or more consecutive characters in the set $\{\vert, \textless, \textgreater, -, /\}$.}. Select the longest chunk as the candidate slogan that either does not contain the company name or has the company name at the beginning (in which case we will strip off the company name and not affect the fluency).
  \item Deduplicate the slogans and keep only the first occurring company if multiple companies have the same slogan.
  \item Filter based on the length of the description and the slogan. The slogan must contain between 3 and 12 words while the description must contain at least 10 words. 
  \item Concatenate the description and the slogan and detect their language using an open-source library~\footnote{\tt{https://github.com/shuyo/language-detection}}. We keep the data only if its detected language is English.
  \item Filter based on lexicographical features, such as the total punctuations in the slogan must not exceed three, the longest word sequence without any punctuation must be at least three words. We come up with these rules based on an analysis of a large number of candidate slogans.
  \item Filter based on named entity tags. We use Stanza~\citep{qi2020stanza} to perform named entity recognition on the candidate slogans. Many candidates contain a long list of locations names. We discard a candidate if over 30\% of its surface text consists of named entities with the tag. ``{\fontfamily{qcr}\selectfont{GPE}}''.
\end{enumerate}

Table~\ref{tab:cleaning-example} provides examples of the cleaning/filtering process. 

\begin{table}[h!]
  \centering
  \caption{Sample descriptions and slogans before and after the data cleaning. Note that ``-'' indicates the algorithm fails to extract a qualified slogan and the example will be removed.}
    \begin{tabular}{p{12.5cm}}
    \hline
    \textbf{Company:} Knorex (\tt{https://www.knorex.com/})\\
   	\textbf{Desc:} Cross-channel marketing cloud platform augmented by Machine Learning. Single dashboard to automate and optimize all your campaigns across digital marketing funnels in one place. \\
   	\textbf{HTML Title:} Cross-Channel Marketing Automation Platform $\vert$ Universal Marketing $\vert$ More Than Just A DSP $\vert$ Knorex.com \\
    \textbf{Extracted Slogan:} Cross-Channel Marketing Automation Platform \\
    \textbf{Explaination:} Longest chunk after splitting; not filtered by any of the rules.
   	\\\hdashline
   	\textbf{Company:} GoPro (\tt{https://gopro.com/en/us/})\\
   	\textbf{Desc:} Discover the world's most versatile action cameras + accessories. Possibilities are endless with waterproof, live streaming, stabilizing features + more. \\
   	\textbf{HTML Title:} GoPro $\vert$ World's Most Versatile Cameras  $\vert$ Shop Now \& Save \\
    \textbf{Extracted Slogan:} World's Most Versatile Cameras \\
    \textbf{Explaination:} Longest chunk after splitting; not filtered by any of the rules.
   	\\\hdashline
    \textbf{Company:} Adfuel Media Inc. (\tt{https://goadfuel.com/}) \\
   	\textbf{Desc:} Adfuel is North America's Digital Marketing agency providing Digital Marketing Services, Programmatic Advertising, Geo-Fencing, and Campaign Budgeting by using a universal advertising platform. \\
   	\textbf{HTML Title:} Digital Marketing Agency in Miami $\vert$ Digital Marketing Agency in Ontario\\
    \textbf{Extracted Slogan:} Digital Marketing Agency in Ontario \\
    \textbf{Explaination:} Longest chunk after splitting; contains one named entity not exceeding 30\% of the surface text.
   	\\\hdashline
    \textbf{Company:} BMW (\tt{https://www.bmw.com/en/index.html})\\
   	\textbf{Desc:} Dive into new worlds with BMW, get inspired, and experience the unknown, the unusual and some useful things, too. \\
   	\textbf{HTML Title:} BMW.com $\vert$ The international BMW Website \\
    \textbf{Extracted Slogan:} - \\
    \textbf{Explaination:} The company name appears in the middle of the candidate slogan ``The international BMW Website'' (Rule 5).\\\hline
    \end{tabular}
  \label{tab:cleaning-example}
\end{table}

\section{Slogan Annotation Guideline for Human Evaluators}
\label{apx:annotation-guideline}

You will be shown five generated slogans for the same company in sequence at each time. They were generated using different models and rules. For each slogan, please rate on a scale of 1-3 (poor, acceptable, good) for each of the three aspects (coherent, well-formed, catchy). Please ensure you rate \textbf{all} the aspects before moving on to the next slogan. Please also ensure your rating standard is consistent both among the candidate slogans for the same company and across different companies. 

Please note that the order of the slogans is randomly shuffled. So you should not use the order information to make a judgement.

The details and instructions are as follows:

\subsection*{Coherent}

A slogan should be coherent with the company description. There are two criteria that it needs to satisfy to be coherent. Firstly, it needs to be \textit{relevant} to the company. E.g., the following slogan is incoherent because there is no apparent link between the description and the generated slogan.

\begin{quote}
Slogan: The best company rated by customers

Description: Knorex is a provider of performance precision marketing solutions
\end{quote}

Secondly, the slogan should not introduce \textit{unsupported} information. E.g., if the description does not mention the company's location, the slogan should not include a location. However, there are cases where the location can be inferred, although the exact location does not appear in the description. We provide some hypothetical examples and the ratings you should provide.

\begin{quote}
Description: Knorex is a provider of performance precision marketing solutions based in \textbf{California}.

Slogan 1: \textbf{US}-based Digital Marketing Company (3, because California infers the company is in the US).

Slogan 2: Digital Marketing Company (3, the slogan does not have to cover all the information in the description).

Slogan 3: Digital Marketing Company in \textbf{Palo Alto} (2, it may be true but we can't verify based on the description alone).

Slogan 4: Digital Marketing Company in \textbf{China} (1, it is false).

\end{quote}

Please focus on verifying \textit{factual} information (location, number, year, etc.) instead of \textit{subjective} description. Expressions like ``Best ...'' or ``Highest-rated ...'' usually do not affect the coherence negatively.

\subsection*{Well-formed}

A slogan should be well-formed, with appropriate \textit{specificity} and \textit{length}. It should also be \textit{grammatical} and \textit{make sense}. Examples that will receive poor scores in this aspect:

\begin{itemize}
  \item Paragraph-like slogans (because it's not concise and inappropriate to form a slogan).
  \item Very short slogans that are unclear what message they convey (e.g., ``Electric Vehicle'').
  \item Slogans containing severe grammatical errors or do not make sense (e.g., slogans that look like a semi-random bag of words).
\end{itemize}

\subsection*{Catchy}

A slogan should be catchy and memorable. Examples are using metaphor, humour, creativity or other rhetorical devices. So slogan A below is better than slogan B (for the company M\&Ms).

\begin{quote}
Slogan A: Melts in Your Mouth, Not in Your Hands

Slogan B: Multi-Coloured Button-Shaped Chocolates
\end{quote}

Lastly, please perform the labelling independently, especially do not discuss with the other annotator performing the same task. Thank you for your contribution!

\section{Annotation Interface}
\label{apx:annotation-ui}

\begin{figure}
  \centering
  \includegraphics[width=\textwidth]{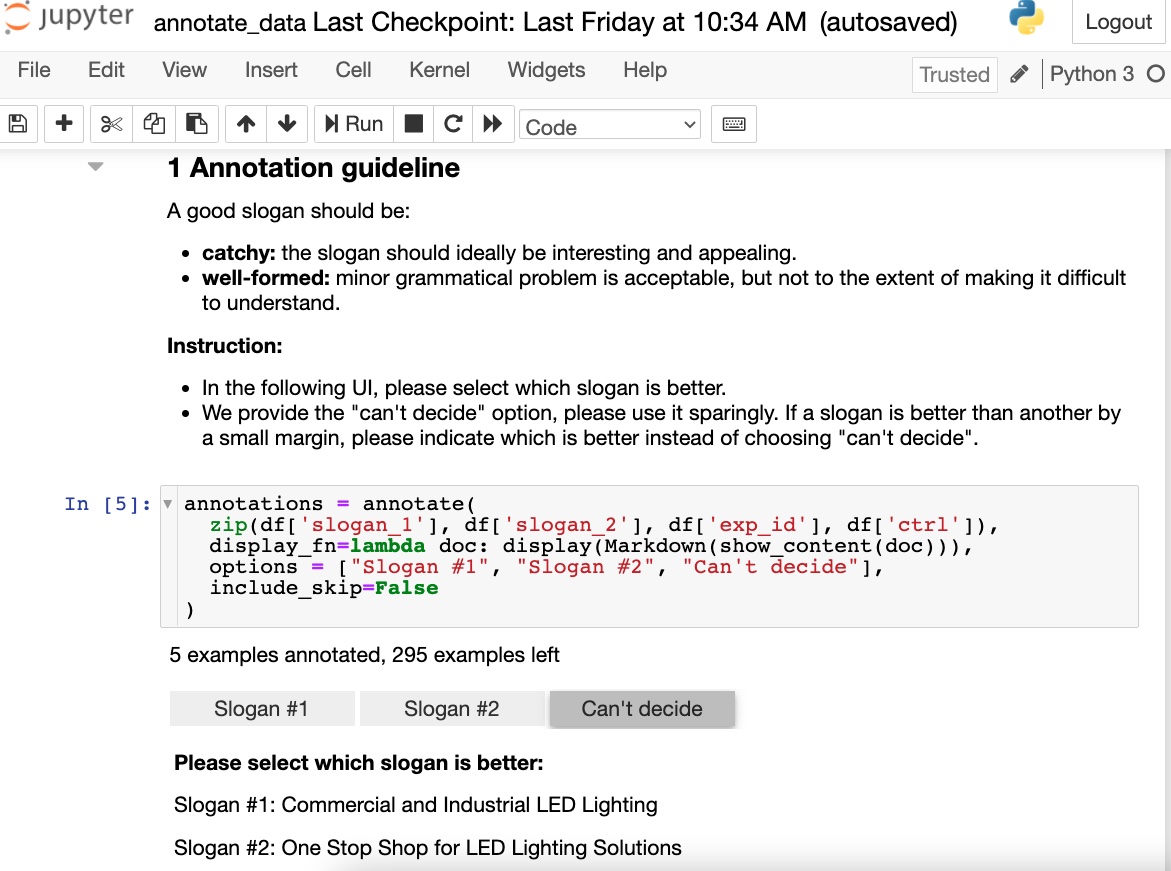}
\caption{User interface for pair-wise slogan ranking described in Section~\ref{subsec:diversity-eval}. One of the candidate slogans uses our proposed syntactic control code, while another candidate uses nucleus sampling. We randomise the order of the slogans to eliminate positional bias.}
\label{fig:diversity_anotation_ui}
\end{figure}

\begin{figure}
  \centering
  \includegraphics[width=\textwidth]{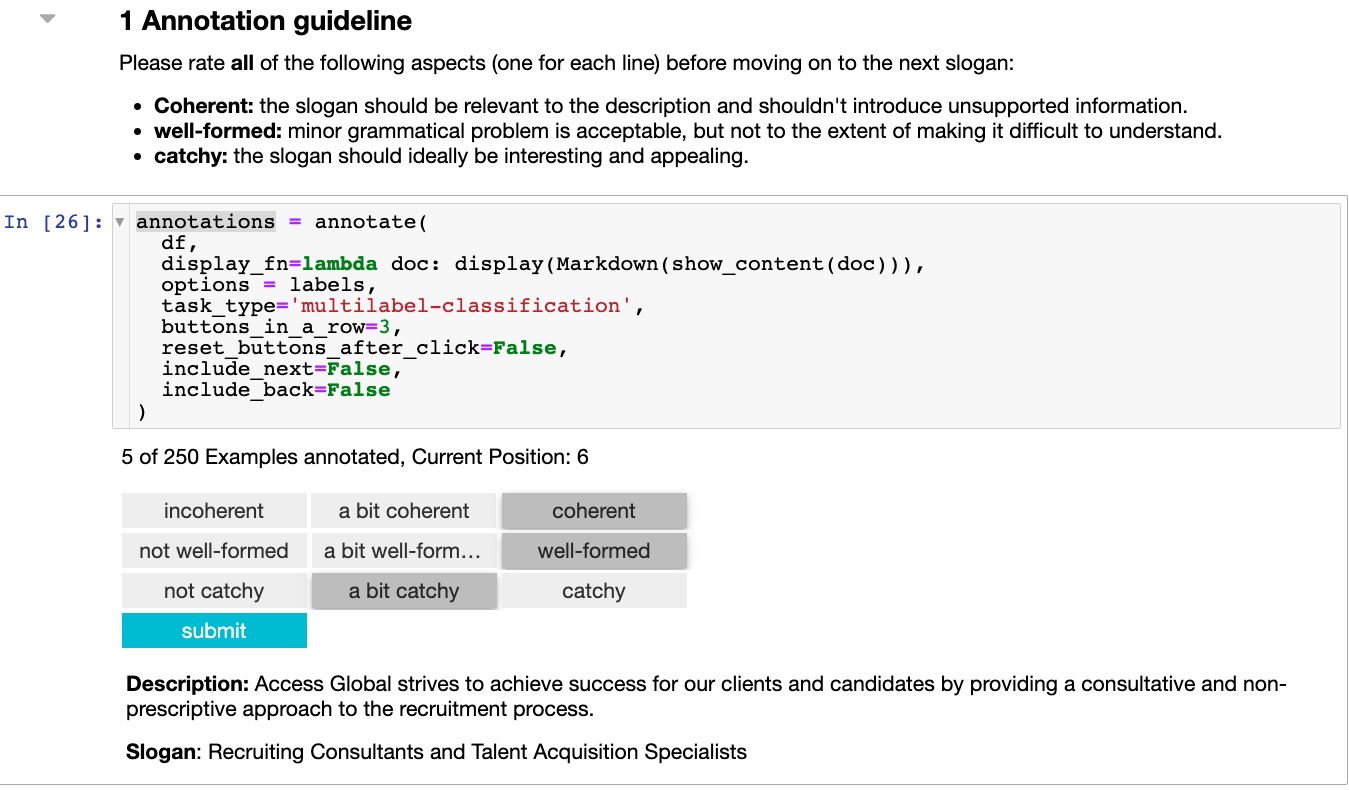}
\caption{User interface for fine-grained slogan evaluation described in Section~\ref{subsec:human-evaluation}. We randomise the order of the slogans to eliminate positional bias.}
\label{fig:final_anotation_ui}
\end{figure}

We implement the human annotation interfaces using Jupyter notebook widgets implemented in Pigeon~\footnote{\tt https://github.com/agermanidis/pigeon}. Figure~\ref{fig:diversity_anotation_ui} shows the UI for pair-wise ranking task conducted in Section~\ref{subsec:diversity-eval} and Figure~\ref{fig:final_anotation_ui} shows the UI for fine-grained evaluation conducted in Section~\ref{subsec:human-evaluation}.

\label{lastpage}

\end{document}